\documentclass{article}

\usepackage{microtype}
\usepackage{multirow}
\usepackage{graphicx}
\usepackage{wrapfig}
\usepackage{subcaption}
\usepackage{bm}
\usepackage{soul}
\usepackage{booktabs} 
\usepackage{multirow}

\usepackage{hyperref}

\usepackage[accepted]{icml2023}

\usepackage{amsmath}
\usepackage{amssymb}
\usepackage{mathtools}
\usepackage{amsthm}
\usepackage{shortcutsg}
\usepackage{yhmath}

\usepackage[capitalize,noabbrev]{cleveref}

\theoremstyle{plain}
\newtheorem{theorem}{Theorem}[]

\newtheorem{corollary}{Corollary}

\theoremstyle{definition}
\newtheorem{definition}{Definition}
\newtheorem{assumption}{Assumption}
\theoremstyle{remark}
\newtheorem{remark}{Remark}
\theoremstyle{result}
\newtheorem{result}{Result}

\usepackage[textsize=tiny]{todonotes}

\newcommand{\x}{\mathbf{x}}
\newcommand{\ti}{\mathrm{t}}
\newcommand{\y}{\mathrm{y}}
\newcommand{\ui}{\mathrm{u}}
\newcommand{\Y}{\mathrm{Y}}
\newcommand{\Yzero}{\Y^0}
\newcommand{\Yone}{\Y^1}
\newcommand{\D}{\mathcal{D}}

\newcommand{\Pf}{P_{\text{full}}}
\newcommand{\Po}{P}
\newcommand{\Q}{Q}
\renewcommand{\E}{\EE}

\icmltitlerunning{B-Learner}

\begin{document}

\twocolumn[
\icmltitle{B-Learner: Quasi-Oracle Bounds on Heterogeneous Causal Effects Under Hidden Confounding}

\icmlsetsymbol{equal}{*}

\begin{icmlauthorlist}
\icmlauthor{Miruna Oprescu}{cornell}
\icmlauthor{Jacob Dorn}{princeton}
\icmlauthor{Marah Ghoummaid}{technion}
\icmlauthor{Andrew Jesson}{oxford}
\icmlauthor{Nathan Kallus}{cornell}
\icmlauthor{Uri Shalit}{technion}
\end{icmlauthorlist}

\icmlaffiliation{oxford}{OATML, University of Oxford}
\icmlaffiliation{technion}{Technion, Israel Institute of Technology}
\icmlaffiliation{princeton}{Princeton University}
\icmlaffiliation{cornell}{Cornell University and Cornell Tech}

\icmlcorrespondingauthor{Miruna Oprescu}{amo78@cornell.edu}

\icmlkeywords{Causal inference, sensitivity analysis, heterogeneous treatment effect, hidden confounding}

\vskip 0.3in
]

\printAffiliationsAndNotice{}  

\begin{abstract}
Estimating heterogeneous treatment effects from observational data is a crucial task across many fields, helping policy and decision-makers take better actions. There has been recent progress on robust and efficient methods for estimating the conditional average treatment effect (CATE) function, but these methods often do not take into account the risk of hidden confounding, which could arbitrarily and unknowingly bias any causal estimate based on observational data. We propose a meta-learner called the B-Learner, which can efficiently learn sharp \emph{bounds} on the CATE function under limits on the level of hidden confounding. We derive the B-Learner by adapting recent results for sharp and valid bounds of the average treatment effect \citep{dorn2021doubly} into the framework given by \citet{kallus2022robust} for robust
and model-agnostic learning of conditional distributional treatment effects. The B-Learner can use any function estimator such as random forests and deep neural networks, and we prove its estimates are valid, sharp, efficient, and have a quasi-oracle property with respect to the constituent estimators under more general conditions than existing methods. Semi-synthetic experimental comparisons validate the theoretical findings, and we use real-world data to demonstrate how the method might be used in practice. 
\end{abstract}

\section{Introduction}

Using data to estimate the causal effect of actions is a fundamental task in medicine, economics, education research, and more. For instance, we might wish to use patient data to estimate which patients react well to a certain medication and which patients should avoid it. In many cases, due to economic and ethical considerations, the data available for these tasks is \emph{observational data}, i.e. data that was not collected as part of a randomized experiment. Using such data carries the risk of \emph{unobserved confounding}: correlations between the observed interventions and outcomes that are not accounted for in the available data. For example, patients with more social support might tend to receive certain interventions over others. If the level of a patient's social support is not recorded in the data, the estimated effect of the intervention will be biased due to not observing the confounder of social support. Unobserved confounding cannot be detected from data, and its presence can lead to arbitrary and unknown bias in causal effect estimates. Such bias can lead to unreliable and potentially harmful policies.

In this work we are concerned with estimating causal effects on an individual level in the presence of a limited degree of unobserved confounding. Specifically, we give a method for effectively learning upper and lower bounds on the conditional average treatment effect (CATE) function that allows for flexible nuisance estimation and high-dimensional conditioning sets like patient medical records. The degree of allowed hidden confounding can be set by domain knowledge; alternatively, we can estimate what degree of hidden confounding is needed to significantly change our understanding of the CATE for any particular instance or sub-population. 

We pursue the desirable treatment effect bound properties of validity, sharpness, efficiency, and robustness.
A bound is called valid if it contains the true value of the causal estimand.
A sharp bound is a valid bound that contains \emph{only} those values of the causal estimand that could emerge from a plausible data generation process that could have produced the observed data \citep{HoAndRosen}.
Therefore, sharp bounds are the smallest possible bounds accounting for both observational data and domain knowledge (in the form of the degree of hidden confounding), a property which is important for precise decision making under hidden confounding.
In contrast, a valid bound could in principle contain extraneous values, leading to overly cautious decision making.
Efficient bounds converge to their target values using as little data as possible. Typically, efficiency at best corresponds to quasi-oracle performance, where only slowly-consistent first-stage estimates are needed to achieve the same error bounds as we would obtain with access to oracle knowledge \citep{nie2021quasi}. Finally, a robust bound will be insensitive (within limits) to biases in the constituent estimators. We formalize these properties in \cref{sec:back}.

In this paper, we present the B-Learner, for ``bound-learner'', a scalable and flexible meta-learner for estimating \emph{bounds} on the CATE function. The B-Learner is a meta-learner that uses a partially double-robust, Neyman-orthogonal estimating equation for the valid CATE bound characterization \citep{dorn2021doubly}. For unconfounded CATE estimation, there are several well-known meta-learners such as the X-Learner \citep{kunzel2019metalearners}, DR-Learner \citep{kennedy2020optimal}, and R-Learner \citep{nie2021quasi}. These methods allow the user to use any combination of learning methods (be it random forest, linear regression, or deep neural nets) and combine them efficiently to estimate the CATE function. In addition to flexibility, some of these methods have desirable rate and quasi-oracle properties. The B-Learner offers analogous flexibility, rate, and quasi-oracle guarantees for CATE \textit{bounds} estimation, as well as novel bound validity and sharpness guarantees under appropriate conditions. We study the CATE bounds under Tan's marginal sensitivity model (MSM) \citep{tan2006distributional}, which quantifies the degree of unobserved confounding through odds ratios. The properties of Tan's MSM give the B-Learner validity under notably weak conditions. 

We evaluate the B-Learner using synthetic and semi-synthetic experiments. In the synthetic experiments, the B-Learner displays quasi-oracle efficiency, requiring only a moderate amount of data for it to perform near-identically with estimated and oracle first-stage nuisances. The B-Learner also performs at least comparably to existing methods with analogous nuisances and can perform better with a well-tailored choice of second-stage regression function. In semi-synthetic experiments, we find the B-Learner is at least as effective as existing state-of-art models on a previously proposed benchmark. Finally, we illustrate the use of the B-Learner using real data to estimate the effect of 401(k) eligibility on financial wealth. 

\textbf{Related work.}
To the best of our knowledge, existing methods for CATE sensitivity analysis have yet to show all four of the properties of our proposed B-Learner.
In particular, \citet{kallus2019intervalEstimation}, \citet{jesson2021quantifying} and \citet{yin2022conformal} each present methods that only achieve validity, and, to some degree, rate properties. These approaches start with estimators that have good properties under unconfoundedness and then optimize the estimated CATE or average treatment effect (ATE) bounds subject to a subset of the constraints implied by Tan's MSM. 
Because these approaches do not impose all implications of the MSM, they lack sharpness outside knife-edge cases. 
They also do not actively exploit the Neyman orthogonality of their solution, so they do not have the same rate guarantees that can be obtained under unconfoundedness. Lastly, these methods are tailored to specific learners and do not allow for the same flexibility as a meta-learner. Works such as \citet{yadlowsky2022bounds} and \citet{chernozhukov2022long} study bounds under other sensitivity assumptions. \citet{yadlowsky2022bounds} exploit Neyman orthogonality to obtain rate guarantees on CATE estimates and fast root-$n$ guarantees on ATE estimates in \citet{RosenbaumObservational}'s model, which they show are sharp when certain outcome symmetry properties hold. 
\citet{chernozhukov2022long} provide a method guaranteeing root-$n$ consistency for \emph{average} potential outcomes, treatment effects, and derivative bounds under limits on variance and covariance, and show that their estimates are sharp provided the bounds do not violate any implications of the observable data distribution.
A rich literature on sensitivity analysis for ATEs exists, from as early as \citet{cornfield1959smoking} through \citet{rosenbaum1983assessing} to recent work like \citet{colnet2022causal}, but commentary on these methods is out of scope for this work.

\section{Background and setup}\label{sec:back}

We work in an observational data setting using the Neyman-Rubin potential outcomes framework. 
We assume data is drawn from an unobservable distribution $\Pf$ over $(X, A, Y(1), Y(0), U)$, where $A \in \{0, 1\}$ is a binary treatment, $X$ is a set of baseline covariates in $\mathbb{R}^d$, $Y(1)$ and $Y(0)$ are the real-valued treated and untreated potential outcomes, respectively, and $U \in \mathbb{R}^k$ is an unobserved confounder. 
However, we face the fundamental problem of causal inference and 
only observe $n$ draws from the coarsened distribution $\Po$ over the observed variables $Z = (X, A, Y )$, where we assume that $Y=Y(A)$, i.e. (causal) consistency.

We are interested in learning about the conditional average treatment effect (CATE):
\begin{align*}
    \tau(x) & = \EE_{\Pf}[Y(1) - Y(0) \mid X=x].
\end{align*} 
The average treatment effect (ATE) is $\EE[\tau(X)]$. When the (untestable) unconfoundedness assumption holds, formally $A \indep Y(1), Y(0) \mid X$, then the CATE is equivalent to the difference in expected observed potential outcomes: $\tau(x) = \EE_{P}[Y \mid X=x, A=1] - \EE_{P}[Y \mid X=x, A=0]$. With the additional assumption of positivity, the CATE can be estimated with standard tools.
However, the unconfoundedness assumption is untestable and often unrealistic, as we often have at least some degree of confounding unaccounted for by the observed covariates $X$. 
Therefore, we will assume unconfoundedness only holds with the addition of an unobserved $U\in \RR^k$ for some $k$, such that $A \indep Y(1), Y(0) \mid X, U$. 
In this case, it is possible to bound $\tau(x)$ pointwise, for example, by assuming that unobserved confounding induces only a limited divergence between $P$ and $\Pf$.

We proceed under Tan's Marginal Sensitivity Model (MSM) \citep{tan2006distributional}. Formally:
\begin{assumption}\label{assumption:msm}
    Let $e(x, u) = \Pf(A = 1  \:|\: X=x, U=u)$ and $e(x) = P(A = 1 \:|\: X=x)$ be the full and observed propensity scores, respectively. 
    We assume $e(x), e(x, u) \in (0, 1)$ and that there exists $\Lambda \geq 1$ such that the following  holds almost surely under $\Pf$:
    \begin{align*} 
        & \Lambda^{-1} \leq \left. \frac{e(x, u)}{1-e(x, u)} \right/ \frac{e(x)}{1-e(x)} \leq \Lambda . 
    \end{align*}
\end{assumption}
The MSM imposes a bound on ratio between the full odds of treatment $e(x, u) / (1-e(x, u))$ and the observed odds of treatment $e(x) / (1-e(x))$. (The MSM is sometimes equivalently described using the log odds ratio bound $\log(\Lambda)$.) When $\Lambda=1$, Assumption \ref{assumption:msm} is equivalent to the classic assumption of unconfoundedness with respect to the observed $X$. As $\Lambda$ grows away from $1$, greater unobserved confounding is allowed under the MSM and we can generally only estimate bounds on the CATE. In this paper our goal is to characterize these bounds, which describe a notion of ``causal'' uncertainty in the CATE estimate. 

\begin{remark} The sensitivity parameter $\Lambda$ is a user-defined hyper-parameter as it specifies how much confounding to allow for. Choosing a suitable $\Lambda$ is an ongoing area of study. \citet{hsu_small2013} propose a procedure where we assess $\Lambda$ values that correspond to dropping observed covariates and using domain knowledge to judge whether we omitted variables as important as these. Inversely, as our intervals increase with $\Lambda$, we can seek the $\Lambda$ where a conclusion or decision would be overturned and judge whether the implied confounding is plausible. Ultimately, the choice of $\Lambda$ is domain-specific and an analyst's choice. 
\end{remark}

\noindent \textbf{Notation.} We now define the main notation, with a more detailed notation table available in \cref{sec:notation}. To unify the analysis for upper (largest plausible CATE) and lower (smallest plausible CATE) bounds, we employ the convention that $+, -$ indicators symbolize upper and lower bounds, respectively. For nuisance functions (e.g. quantiles), these signs also encode the dependence on $\alpha = \Lambda / (\Lambda+1)$ (and $\Lambda$) which we otherwise generally suppress in the remainder of the paper.  
We define the conditional outcome quantile and shorthand quantile notation:
\begin{align*}
    q_{c}^*(x, a) & = \inf\{\beta: F(\beta\mid x, a)\geq c\} \\
    q_+^*(x, a) & =  q_\alpha^*(x, a), q_-^*(x, a) =  q_{1-\alpha}^*(x, a). 
\end{align*}
The $\pm$ and $\mp$ symbols signal that an equation should be read twice, once with $\pm=+, \mp=-$ and once with $\pm=-, \mp=+$ (see example in \cref{sec:notation}, \cref{tab:notation}).  
For conciseness and clarity, we focus our main discussion on CATE upper bounds. In \cref{sec:lbResults}, we provide a similar analysis of CATE lower bounds.

\subsection{Properties of bound estimates} \label{sec:bound-props}

Our goal is to estimate the \emph{identified set}: the set of CATEs that can be obtained in the unobserved distribution $\Pf$ generating the observed distribution $\Po$ and satisfying the requirements of \cref{assumption:msm}. 

\begin{definition}\label{def:IdentifiedSet}
    The \emph{identified set} of estimands under \cref{assumption:msm} is the set of estimands that can be obtained for a distribution $\Q$ over $(X, A, Y(1), Y(0), U)$ 
    such that the distribution of $(X, A, Y)$ under $\Q$ matches the observed distribution $\Po$ and $\Lambda^{-1} \leq \left. \frac{\Q(A=1 \mid X=x, U=u)}{\Q(A=0 \mid X=x, U=u)} \right/ \frac{e^*(x)}{1-e^*(x)} \leq \Lambda$ almost surely. 
    Let $\mathcal{M}(\Lambda)$ be the set of distributions $Q$ that the observed data $Z=(X, Y, A)$ and \cref{assumption:msm} cannot rule out. Then, the \emph{sharp} (upper) bounds on the identified set of conditional average potential outcomes and CATEs for a given point $x$ are given by:
    \begin{align*}
        Y^{+}(x, a) & \equiv \sup_{\Q \in \mathcal{M}(\Lambda)} \E_{\Q}[ Y(a) \mid X=x]  \\
        \tau^{+}(x) & \equiv \sup_{\Q \in \mathcal{M}(\Lambda)} \E_{\Q}[ Y(1) - Y(0)  \mid X=x] .
    \end{align*} 
\end{definition} 
Lower bounds follow symmetrically by replacing the suprema with infima.
We note that the requirements of \cref{assumption:msm} decouple across $x$ and are convex, so finding the identified set reduces to finding pointwise bounds.
As we will see in \cref{sec:bounds-id}, the CATE upper bounds $\tau^{+}(x)$ depend only on the observed distribution of data $Z$ and the sensitivity parameter $\Lambda$. We can therefore ask what good properties we might want estimates $\widehat{\tau}^{+}(x)$ to have. We suggest four desirable properties for bound estimation, of which the last two are closely linked: 

\begin{figure*}[ht!]
     \centering
     \begin{subfigure}[b]{0.5\textwidth}
         \centering
         \includegraphics[width=\textwidth]{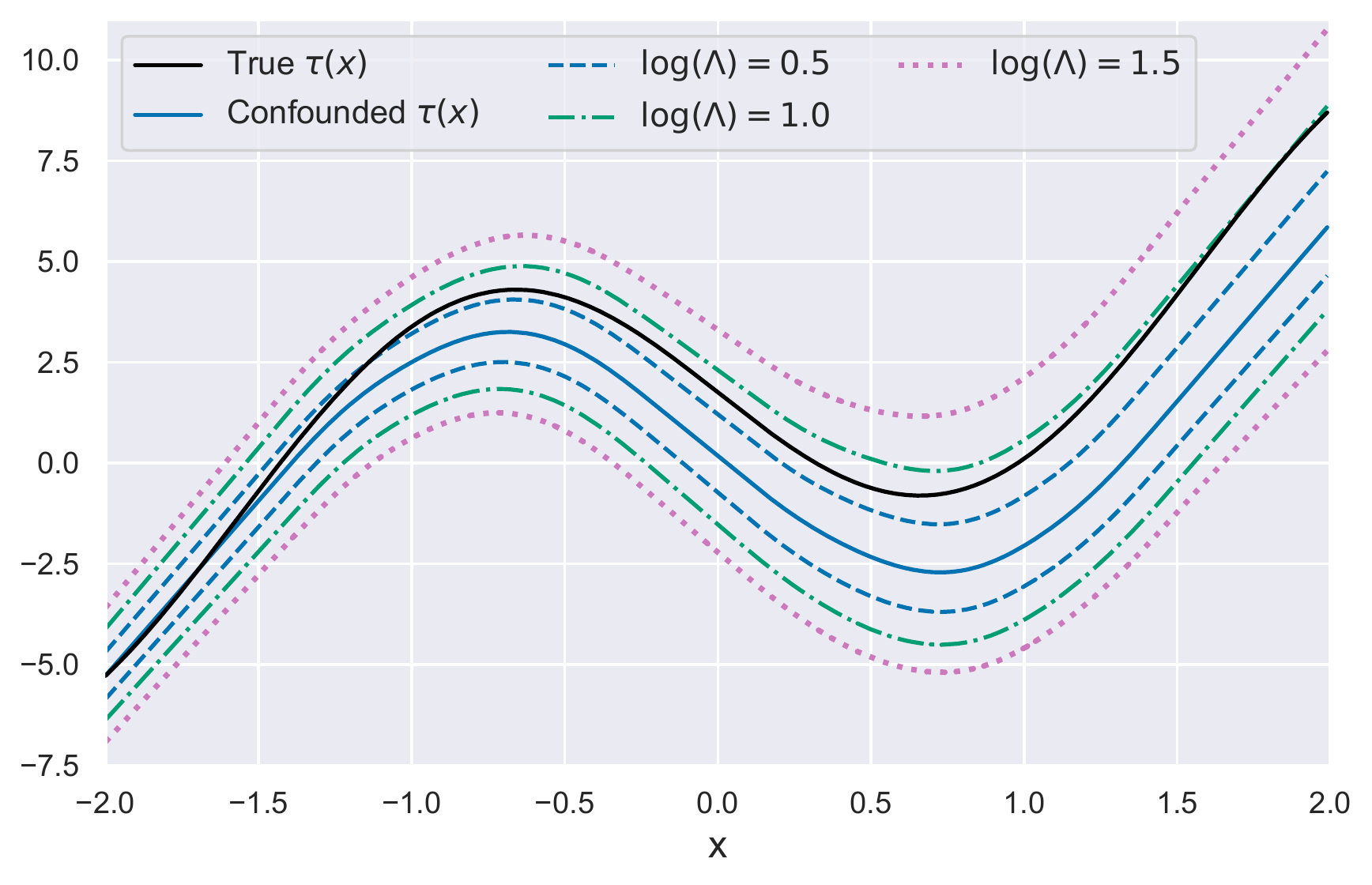}
         \caption{CATE bounds for different values of $\Lambda$}
         \label{fig:lambda-bounds}
     \end{subfigure}
     \begin{subfigure}[b]{0.5\textwidth}
         \centering
         \includegraphics[width=\textwidth]{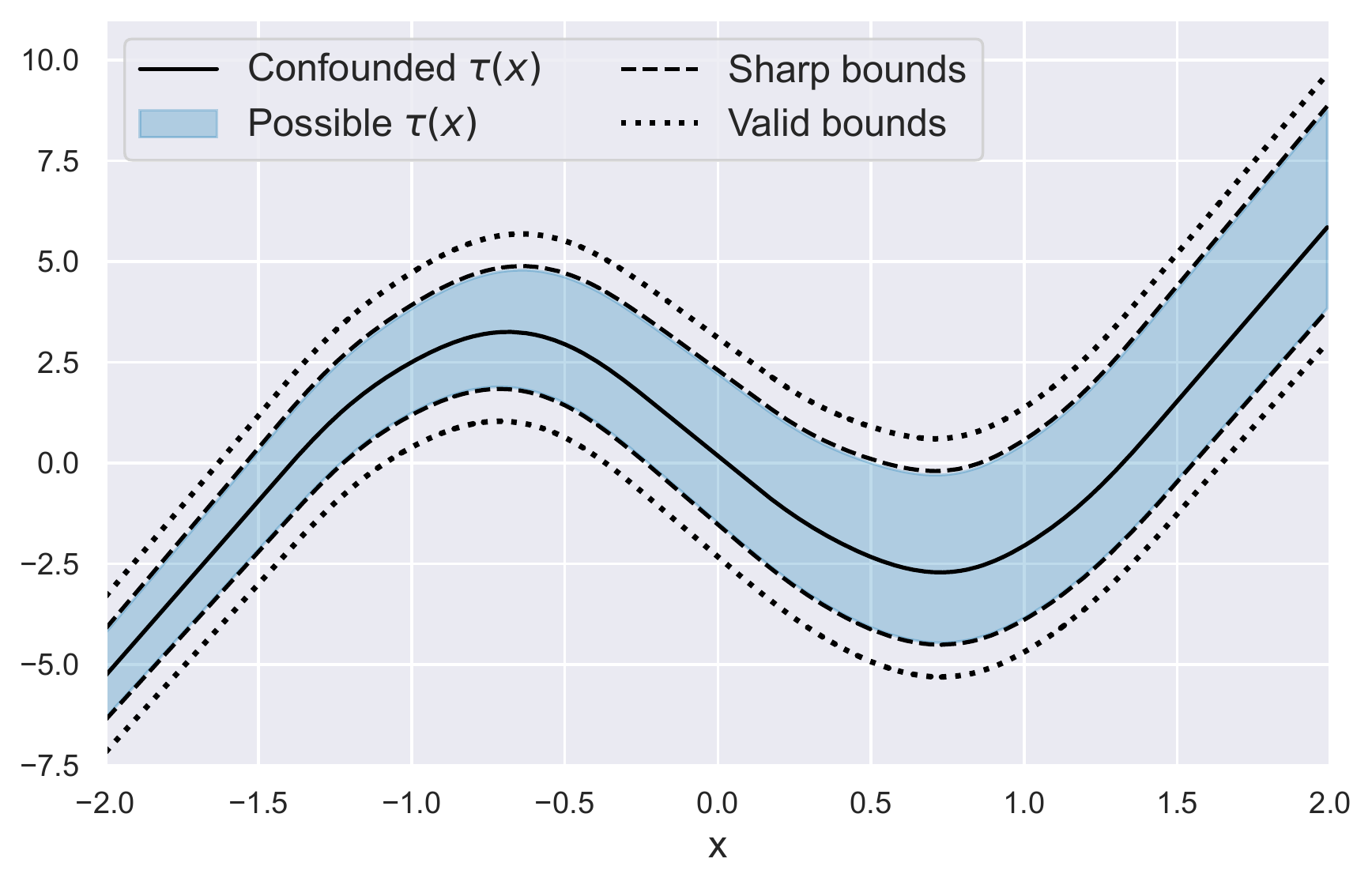}
         \caption{Sharp and valid bounds}
         \label{fig:sharp-valid-bounds}
     \end{subfigure}
        \caption{Example of a confounded CATE function with a true odds ratio $\Lambda^*$ given by $\log(\Lambda^*)=1.0$. The true $\tau(x)$ is the unobserved CATE in the full distribution, $\EE_{\Pf}[Y(1) - Y(0) \mid X=x]$. The confounded $\tau(x)$ is the biased estimand under assumed uncondoundedness, $\EE_{P}[Y\mid X=x, A=1] - \EE_{P}[Y\mid X=x, A=0]$.
        (\ref{fig:lambda-bounds}): The sharp bounds from \cref{result:sharp-cate-bounds} for different levels of $\Lambda$. The true $\tau(x)$ is inside the sharp bounds for $\log(\Lambda) = 1.0$ (green). (\ref{fig:sharp-valid-bounds}): Example of \textit{valid} and \textit{sharp} CATE bounds as defined in \cref{sec:bound-props}.}
        \label{fig:bound-examples}
     \vspace{-0.5em}
\end{figure*}

\textbf{Valid estimates.} If $\widehat{\tau}^{+}(x) < \tau^{+}(x) - o_p(1)$, then our estimated bounds would 
fail to cover the identified set and rule out plausible CATEs even asymptotically, which would be undesirable. Conversely, bound characterizations $\bar{\tau}$ satisfying $\bar{\tau}^{+}(x) \geq \tau^{+} (x)$ are called ``valid" in the partial identification literature \cite{HoAndRosen}, since Assumption \ref{assumption:msm} implies $\tau^{+} (x) \geq \E[Y(1) - Y(0) \mid X=x]$. Valid bounds (illustrated in \cref{fig:bound-examples}) give us some but not all information from our assumptions: every value they rule out is implausible, but some values they do not rule out may be implausible as well. We relax the notation and say that bound estimates $\widehat{\tau}$ are \emph{valid} if $\widehat{\tau}^{+}(x) \geq \tau^{+}(x) - o_p(1)$. 

\textbf{Sharp estimates.} If $\widehat{\tau}^{+}(x) > \tau^{+}(x) + o_p(1)$, then our estimated bounds would fail to rule out impossible CATEs asymptotically under our assumptions. 
Exact characterizations of the identified sets are called ``sharp'' in the partial identification literature \cite{HoAndRosen}. Sharpness is a stronger property than validity. We use lax notation to say that bound estimates $\widehat{\tau}$ are \emph{sharp} if $\widehat{\tau}^{+}(x) = \tau^{+}(x) + o_p(1)$. 

\textbf{Efficient and robust estimates.} We would like our bound estimates to converge to their limits at desirable rates and have multiple chances at sharp or valid limits. Ideally, we would be able to learn CATE bounds at the same rate as we could obtain under unconfoundedness. These properties relate to ``double robust" estimators and may require constructing Neyman-orthogonal characterizations of valid, and ideally sharp, bounds. 

\subsection{Identification and estimation of sharp bounds}\label{sec:bounds-id}

In this section, we use results from \citet{dorn2021doubly} to show how we can identify and estimate sharp CATE bounds from the observed data distribution, $P$. 
In order to express the sharp bounds, we introduce the following pseudo-outcomes from \citet{dorn2021doubly} that will correspond to the Conditional Value at Risk and the unobserved outcome bounds under Assumption \ref{assumption:msm}:
\begin{align*}
    H_{\pm}(z, \bar{q}) & = \bar{q}(x, a) + \frac{1}{1-\alpha} \left\{y-\bar{q}(x, a)\right\}_{\pm} \\
    R_\pm(z, \bar{q}) &= \Lambda^{-1}y+(1-\Lambda^{-1}) H_{\pm}(z, \bar{q}) \\
    \rho^*_{\pm}(x, a, \bar{q}) &= \EE[ R_{\pm}(z, \bar{q})\mid X=x, A=a]. 
\end{align*}
We use the shorthand $\rho_\pm^*(x, a)=\rho^*_{\pm}(x, a, q_\pm^*)$ to write the $\rho^*_{\pm}$ function evaluated at the true conditional quantiles $q^*_{\pm}$ (which will end up corresponding to sharp bounds). 
The quantity $\text{CVaR}_{\pm}(x, a):=\EE[ H_{\pm}(z, q_\pm^*)\mid X=x, A=a]$ is known as the Conditional Value at Risk \citep{artzner1999coherent, kallus2022treatment}. In the distribution  $Y\mid X=x, A=a$, $\text{CVaR}_+(x, a)$ is the expectation above the $(1-\alpha)$ quantile, whereas  $\text{CVaR}_-(x,a)$, is the expectation below the $\alpha$ quantile. Hence, the pseudo-outcomes $H$ and $R$ correspond to the Conditional Value at Risk  
and conditional unobserved potential outcome, respectively. 

Let $\mu^*(x, a) = \E[Y \mid X=x, A=a]$ be the conditional outcome regression in the observed data.
Note that we can write the conditional potential outcome under $Q$ 
as $\E_{\Q}[ Y(a) \mid X=x] = \Po[A=a \mid X=x] \mu^*(x, a) + \Po[A =1- a \mid X=x] \E_Q[Y(1-a)\mid X=x, A=a]$ since $Q$ must be consistent with the observed distribution $P$. Thus, it suffices to bound the conditional unobserved 
potential outcome $\E_Q[Y(1-a)\mid X=x, A=a]$, which leads to the following result in terms of $\rho_{\pm}^*(x, a) = \rho_{\pm}^*(x, a, q_{\pm}^*)$:

\begin{result}[Sharp bounds, \cite{dorn2021doubly}]\label{result:sharp-cate-bounds} 
    The conditional average unobserved potential outcome $\E_Q[Y(1-a)\mid X=x, A=a]$ has sharp upper and lower bounds under Assumption \ref{assumption:msm} given by $\rho^*_+(x,a)$ and $\rho^*_-(x,a)$, respectively. Thus, the sharp bounds on the conditional average potential outcomes can be written as: 
    \begin{align*}
        Y^{+}(x, 1) &= e^*(x) \mu^*(x, 1) + (1-e^*(x)) \rho_{+}^*(x, 1)\\
        Y^{-}(x, 0) &= (1-e^*(x)) \mu^*(x, 0) + e^*(x) \rho_{-}^*(x, 0).
    \end{align*}
    The sharp CATE upper bound is further given by $\tau^{+}(x) = Y^{+}(x, 1) - Y^{-}(x, 0)$. 
\end{result}
Thus, the bounds are a convex combination of the conditional outcome function $\mu^*(x, a)$ and the corresponding conditional CVaR terms, all of which can be estimated from $\Po$. As $\Lambda$ grows, both the weight on the CVaR term in $\rho^*$ grows and the CVaR term itself become more extreme. If the wrong putative quantile $\bar{q}$ is used instead of the true $q^*$, the CVaR term moves the bound in a conservative yet valid direction. Finally, the difference between sharp conditional average potential outcome bounds $\tau^{+}(x)$ clearly yields valid CATE bounds; those bounds are shown to be sharp by arguments outside the scope of this paper \citep{DornAndGuoQuantileBalancing}. As we will see, this characterization of sharp and valid bounds alone will be insufficient for quasi-oracle estimation. 

\textbf{Pseudo-outcome regression for quasi-oracle estimation.} The expression of $\tau^{+}(x)$ suggests a plug-in strategy. We can estimate $e$, $\mu$, and $\rho$ through classification and regression and obtain bound estimates. However, such plug-in estimators are known to suffer from excessive bias due to the estimated nuisances \cite{kennedy2020optimal, kallus2022robust}, especially when the nuisance functions are more complex than the CATE bounds. We follow the \citet{kallus2022robust} strategy and derive an efficient pseudo-outcome for the bounds based on the relevant influence function; we then regress that pseudo-outcome on $X$. We build on this literature to similarly provide an estimator for sharp CATE bounds with desirable properties beyond those of the plug-in estimators implied by \cref{result:sharp-cate-bounds}.

Henceforth we will refer to $e, q, \rho$ as nuisances as they will need to be estimated from data.

\section{B-Learner: Pseudo-Outcome Regression for Doubly-Robust Sharp CATE Bounds}

We propose a debiased learning procedure that consists of regressing a carefully constructed and nuisance-debiasing pseudo-outcome on covariates \cite{kennedy2020optimal, kallus2022robust}. 

\begin{definition}[CATE Bounds Pseudo-Outcome]\label{def:pseudooutcome}
Let $\widehat{\eta} = (\widehat{e}, \widehat{q}_{-}(\cdot, 0), \widehat{q}_{+}(\cdot, 1), \widehat{\rho}_{-}(\cdot, 0), \widehat{\rho}_{+}(\cdot, 1))\in \Xi$ be a set of estimated nuisances. We define the pseudo-outcome corresponding to the bounds for $Y^+(x, 1), Y^-(x,0)$ and $\tau^{+}(x)$ from Result \ref{result:sharp-cate-bounds} by: 
\begin{align*}
        & \phi^{+}_1(Z, \widehat{\eta}) =  AY + (1-A)\widehat{\rho}_{+}(X, 1) \\
        & \qquad \ts+\frac{\left(1-\widehat{e}(X)\right)A}{\widehat{e}(X)}\cdot \left(R_{+}(Z, \widehat{q}_{+}(X, 1))-\widehat{\rho}_{+}(X, 1)\right),\\
         & \phi^{-}_0(Z, \widehat{\eta}) =  (1-A)Y + A\widehat{\rho}_{-}(X, 0) \\
        & \qquad + \ts\frac{\widehat{e}(X)(1-A)}{\left(1-\widehat{e}(X)\right)}\cdot \left(R_{-}(Z, \widehat{q}_{-}(X, 0))-\widehat{\rho}_{-}(X, 0)\right),
    \\
    &\phi^{+}_{\tau}(Z, \widehat{\eta}) =  
     \phi^{+}_1(Z, \widehat{\eta}) - \phi^{-}_0(Z, \widehat{\eta}).
\end{align*}
\end{definition} 

The expressions in \cref{def:pseudooutcome} depend purely on the observed data distribution $\Po$, and so can be viewed as statistical estimands to be learned from the observed distribution. 

When $\Lambda=1$ and unconfoundedness holds, the expression for $\phi^{+}_{\tau}(Z, \widehat{\eta})$ reduces to the familiar doubly-robust pseudo-outcome for CATE estimation, $\widehat{\mu}(X, 1)-\widehat{\mu}(X, 0)+\frac{A-\widehat{e}(X)}{\widehat{e}(X)(1-\widehat{e}(X))} (Y-\widehat{\mu}(X, A))$ \citep{kennedy2020optimal,knaus2022double}.
    
The pseudo-outcome is based on the efficient influence function of the estimand $\EE[\tau^{+}(X)]$, so as we will see, small errors in the nuisance estimation lead to ``doubly small'' (second-order) errors in the $\widehat{\tau}^{+}(x)$ estimates. 
This special structure orthogonalizes the $\widehat{\rho}_{+}$ estimation error in the plug-in bound estimand $A Y + (1-A) \widehat{\rho}_{+}$ using the added term $\frac{(1 - \widehat{e}(X)) A}{\widehat{e}(X)}( R_{+} - \widehat{\rho}_{+})$ that debiases $\widehat{\rho}_{+}$ estimation error. The weighted CVaR terms $\rho_{\pm}^*(X, a, \bar{q})$ involve an objective which is sharpest when $\bar{q}_{\pm} = q_{\pm}^*$ and which turns out to have a second-order dependence on $\bar{q}_{\pm} - q_{\pm}^*$. Thus, quantile regression errors will move the pseudo-outcome in a conservative but still valid direction and consistent quantile regression errors will have favorable rate properties. 

\begin{algorithm}[ht!]
    \caption{The B-Learner (detailed in \cref{sec:detailed-alg})}\label{alg:sharp-cate}
    \begin{algorithmic}[1]
        \INPUT Data $\{(X_i, A_i, Y_i): i\in \{1,...,n\}\}$, folds $K\geq 2$, nuisance estimators, regression learner $\widehat{\EE}_n$
         \FOR{$k\in \{1,...,K\}$} 
          \STATE Use data $\{(X_i, A_i, Y_i): i\neq k-1\;(\text{mod}\; K)\}$ to construct nuisance estimates $\widehat{\eta}^{(k)}=(\widehat e^{(k)},\widehat q^{(k)},\widehat \rho^{(k)})$
            \FOR{$i= k-1\;(\text{mod}\; K)$} 
                \STATE Set $\widehat{\phi}^{+}_{\tau, i}=\phi^{+}_{\tau}(Z_i, \widehat{\eta}^{(k)})$
        \ENDFOR
        \ENDFOR
        \OUTPUT $\widehat{\tau}^{+}(x)=\widehat{\EE}_n[\widehat{\phi}^{+}_{\tau}\mid X=x]$
    \end{algorithmic}
\end{algorithm}
\vspace{-0.3em}

\textbf{B-Learner.} We call our full two-stage estimation procedure the \emph{B-Learner}. Our procedure is summarized in \cref{alg:sharp-cate} (see \cref{sec:detailed-alg} for a detailed version). In the first stage, we estimate the nuisances (outcome regression, propensity score, CVaR) with $K$-fold cross-fitting and construct Neyman-Orthogonal pseudo-outcome estimates based on \cref{def:pseudooutcome}. In the second stage, we regress the estimated pseudo-outcomes on our covariates $X$, resulting in an estimated CATE bound function.
As we will see now, the properties of this function depend on both the choice of nuisance estimators and the second-stage regressor. 

\textbf{Nuisance estimation.} The propensity score $e^*(x)$ can be estimated using any standard probabilistic binary classifier. The quantiles $q_\pm^*$ can be likewise estimated using any of several standard quantile regression methods \cite{yu1998local, meinshausen2006quantile, athey2019generalized}. The modified outcome regression $\rho_\pm^*(x, a) = \Lambda^{-1} \mu^*(x, a) + (1-\Lambda)^{-1} CVaR_{\pm}(x, a)$ is less standard, but it can be learned by either treating the CVaR pseudo-outcome $R_\pm$ as an outcome, or separately learning the $\mu^*$ and $\text{CVaR}_\pm$ components of $\E[R_\pm \mid X=x, A=a]$. In the first approach, where we plug in the estimated quantiles into the expression for $R_\pm(Z, \bar{q})$ and then regress $R_\pm$ onto $X$ using any standard regressor, further sample splitting is theoretically required for estimating $q^*$ and $\rho^*$. In the second approach, we can learn the $\mu^*$ and $\text{CVaR}$ components on the same sample and then weight them accordingly to obtain estimates of $\rho^*$. The outcome regression $\mu^*(x, a)$ can be estimated via any regression learner and $\text{CVaR}_\pm$ can be likewise estimated using several existing approaches \cite{athey2019generalized,kallus2022robust}.

\section{Theoretical properties of the B-Learner}

We now describe the theoretical properties of our estimator. All proofs are in \cref{sec:proofs}. In \cref{sec:TheoryPseudoOutcome}, we use \citet{kallus2022robust}'s generic approach and \citet{dorn2021doubly}'s validity results to study the bias of the pseudo-outcome with first-stage nuisances.
The pointwise bias from the \textbf{sharp} bounds is on the order of $|\widehat{e} - e^*| |\widehat{\rho} - \rho^*| + (\widehat{q} - q^*)^2$.
When the quantiles are inconsistent, $(\widehat{q}_{+} - q_{+}^*)^2$ and $(\widehat{q}_{-} - q_{-}^*)^2$ do not vanish. The pseudo-outcome bounds still remain \textbf{valid} in expectation, and any bias in the direction of failing to cover the identified CATE set disappears at a rate on the order of $|\widehat{e} - e^*| |\widehat{\rho} - \rho^*(\cdot, \widehat{q})|$. 
In \cref{sec:TheoryERM}, we 
characterize the second-stage regression and we show that we can learn CATE bounds at a \textbf{rate} dominated by the complexity of the target class. As a result, the estimator has robustness properties from the product-of-errors bias, with two chances at \textbf{sharp} bounds in $L_2$ norm and two chances at \textbf{valid} bounds on average. Our main text focuses on ERM-based second stage estimators with $L_2$ sharp bound guarantees. We show similar guarantees hold pointwise for linear smoother second-stage estimators in \cref{sec:LinearSmoothers}.

\subsection{Pseudo-outcome properties}\label{sec:TheoryPseudoOutcome}

We first analyze the bias in our proposed pseudo-outcomes. 
\begin{definition}[Conditional Pseudo-outcome Bias]\label{def:CondPOBias} Take $\widehat{\eta} \in \Xi$ be a set of estimated nuisances and let $\diamond\in\{0, 1, \tau\}$. We define the signed conditional pseudo-outcome bias:  $\mathcal{E}^+_{\diamond}(x; \widehat{\eta}) = \EE\left[\phi^{+}_\diamond(Z, \widehat{\eta}) - \phi^{+}_\diamond(Z, \eta^*)\mid X=x\right]$ and $\mathcal{E}^-_{\diamond}(x; \widehat{\eta}) = \EE\left[\phi^{-}_\diamond(Z, \widehat{\eta}) 
- \phi^{-}_\diamond(Z, \eta^*) \mid X=x\right]$.
\end{definition} 
It immediately follows from \cref{def:CondPOBias} that $\mathcal{E}^+_{\tau}(x; \widehat{\eta}) = \mathcal{E}^{+}_1(x; \widehat{\eta}) - \mathcal{E}^{-}_0(x; \widehat{\eta})$ and $|\mathcal{E}^+_{\tau}(x; \widehat{\eta})| \leq |\mathcal{E}^{+}_1(x; \widehat{\eta})| + |\mathcal{E}^{-}_0(x; \widehat{\eta})|$. 
The pseudo-outcome bias can be understood as the error incurred when performing pseudo-outcome regression with estimated nuisances rather than oracle nuisances. While any bias is undesirable, bias in one direction is worse. When $\mathcal{E}_{\tau} > 0$, the pseudo-outcomes are biased in a conservative but still valid direction. When $\mathcal{E}_{\tau} < 0$, the expected pseudo-outcomes are too aggressive and in expectation exclude plausible CATEs. 

Our pseudo-outcomes fit into the framework of \citet{kallus2022robust} since the estimands and the nuisances are the solutions of conditional moment restrictions (see Proof of \cref{thm:condneym}). Thus, under mild boundedness conditions, we can leverage their results to upper bound $|\mathcal{E}^+_\diamond|$.
\begin{assumption}[Boundedness]\label{asm:bounded} Let $\widehat{\eta}\in \Xi$ be a set of estimated nuisances, and take $\overline{\eta}\in \text{conv}\{(\eta^*, \widehat{\eta})\}$. 
\\(i) $P(\epsilon \leq e^*(x), \widehat{e}(x) \leq 1-\epsilon)=1$ for some $\epsilon>0$. \\ (ii) 
$Y, \overline{q}_{+}(\cdot, 1),  \overline{q}_{-}(\cdot, 0), \overline{\rho}_{+}(\cdot, 1), \overline{\rho}_{-}(\cdot, 0), f(\overline{q}_{+}(x, 1)\;|\; x, 1)$, $f(\overline{q}_{-}(x, 0) \mid x, 0)$ are all uniformly bounded. 
\end{assumption}
The first condition in \cref{asm:bounded} is a standard requirement known as positivity, ensuring that both treatments and controls can be observed for any $X$ with non-zero probability. The second condition is a common boundedness assumption often made in debiased machine learning for ATE and CATE in order to control the growth of $|\mathcal{E}^+_\tau|$. We now state the conditional Neyman orthogonality result we require, which we derive using the tools from \citet{kallus2022robust} and \citet{dorn2021doubly}.

\begin{theorem}[Pseudo-Outcome Conditional Neyman Orthogonality]\label{thm:condneym}
Suppose \cref{asm:bounded} holds. Then a Neyman-orthogonal characterization of the conditional outcome moment $\EE[A Y + (1-A) \rho_{+}^*(X, 1) - Y^{+}(X, 1) \mid X] = 0$ has the form of $\phi_1^{+}$ from \cref{def:pseudooutcome}, and the symmetric result holds for $\phi_0^{-}$. The absolute bias of the CATE upper bound has the product of rates bound: 
\begin{align*}
    \left| \mathcal{E}^{+}_\tau(x; \widehat{\eta}) \right|  &\lesssim |\widehat{e}(x)-e^*(x)|\;|\widehat{\rho}_{+}(x, 1)-\rho_{+}^*(x, 1)| \\ 
    & \quad + |\widehat{e}(x)-e^*(x)|\;|\widehat{\rho}_{-}(x, 0)-\rho_{-}^*(x, 0)| \\
    & \quad + (\widehat{q}_{+}(x, 1) - q_{+}^*(x, 1))^2 \\
    & \quad + (\widehat{q}_{-}(x, 0) - q_{-}^*(x, 0))^2.
\end{align*}
The undesirable direction of bias has the more favorable bound in terms of $\rho^*(x, a, \widehat{q})$: 
\begin{align*}
    \mathcal{E}^+_\tau(x; \widehat{\eta}) & \gtrsim -|\widehat{e}(x)-e^*(x)|\;|\widehat{\rho}_{+}(x, 1)-\rho_{+}^*(x, 1, \widehat{q}_{+})| \\ 
    & \quad - |\widehat{e}(x)-e^*(x)|\;|\widehat{\rho}_{-}(x, 0)-\rho_{-}^*(x, 0, \widehat{q}_{-})|. 
 \end{align*} 
\end{theorem} \vspace{-0.5em}
\cref{thm:condneym} lets us characterize the pseudo-outcome biases.

\textbf{Sharp pseudo-outcome bias.} 
An immediate results is that the pseudo-outcome bias for the CATE (upper) bound is pointwise ``doubly sharp" \cite{dorn2021doubly}: its bias tends to zero if $\widehat{q}_{\pm}$ and one of $\widehat{e}$ or $\widehat{\rho}_{\pm}$ are consistent, and the \textbf{rate} of bias goes to zero faster than the individual nuisances if all nuisances are consistent.

\textbf{Valid pseudo-outcome bias.} In some cases it may be difficult to estimate quantiles consistently or at a sufficient rate for the quantile error $(\widehat{q} - q^*)^2$ to vanish faster than $|\widehat{e}-e^*|\;|\widehat{\rho} - \rho^*|$. If so, the absolute value of pseudo-outcome bias relative to sharp bounds might be relevant to the second-stage estimates, but the level of bias in the direction of failing to cover the identified set still disappears at a product rate $|\widehat{e}-e^*|\;|\widehat{\rho} - \rho^*(\cdot, \widehat{q})| $. The pseudo-outcome estimator is therefore ``doubly valid" \cite{dorn2021doubly}: its undesirable bias tends to zero if one of $\widehat{e}$ or $\widehat{\rho}_{\pm}$ is consistent, and the \textbf{rate} of bias goes to zero faster than the individual nuisances if both are consistent. 

Next, we leverage these results to illustrate the quasi-oracle properties of our B-Learner.

\subsection{ERM-based estimators}\label{sec:TheoryERM}

We consider \cref{alg:sharp-cate} with an empirical risk minimization (ERM) algorithm as the second-stage estimator. In other words, given a class of functions $\mathcal{F}\subset [\Xcal\rightarrow \RR]$, the regression learner $\widehat{\EE}_n$ satisfies:
\begin{equation}\label{eq:erm}
    \widehat{\EE}_n\left[\widehat{\phi}^{+}_{\tau}\mid X=\cdot\right] \in \arg\min_{f\in\Fcal} \frac{1}{n}\sum_{i=1}^n  \left(\widehat{\phi}^{+}_{\tau, i}-f(X_i)\right)^2.
\end{equation}
In this scenario, the error rates of our estimation procedure depend on the complexity of the class $\Fcal$. These were studied in the context of learning with nuisance components in several works including \citet{foster2019orthogonal, kallus2022robust}. The implication of \cref{thm:condneym} is we can immediately apply \citet{kallus2022robust}'s Theorem 2 in our setting, employing bracketing entropy as a class complexity measure. 
We note that bracketing entropy is a \textit{global} technique, with guarantees on the $L_2$ loss over the support of the estimand, in contrast with the \textit{local} methods presented in \cref{sec:LinearSmoothers} which enable pointwise guarantees. 

\begin{corollary}[Rates for ERM Estimators, Theorem 2 from \citet{kallus2022robust}]\label{thm:erm}
Suppose \cref{asm:bounded} holds for $\widehat{\eta}^{(k)}\in\Xi, k\in\{1,...,K\}$.
Let $\mathcal{E}^+_\tau(x):=\sum_{k=1}^K\mathcal{E}_\tau^+(x;\widehat{\eta}^{(k)})$ and let $\widehat{\EE}_n[\cdot\mid X=x]$ be as in \cref{eq:erm}. Further, suppose $\Fcal$ is convex and closed and has bracketing entropy $\log N_{[]}(\Fcal,\epsilon)\lesssim \epsilon^{-r}$ with $0<r<2$ and that $|f(x)|$ is bounded $\;\forall f\in\Fcal,x\in\Xcal$. Then,
\begin{align*}
\|\widehat{\tau}^{+}(x)-\tau^{+}(x)\|  \lesssim O_p(n^{-1/(2+r)})+\|\mathcal{E}^+_{\tau}(x)\|.
\end{align*}
\end{corollary}

\textbf{Second-stage sharp consistency and robustness.} When $\|\mathcal{E}^+_{\tau}(x)\| = o_p\left( 1 \right)$ and the conditions above hold, \cref{thm:erm} shows that ERM estimates are $L_2$ consistent for the sharp CATE bounds. Learners satisfying the conditions of \cref{thm:erm} include sparse linear models, neural networks, kernel classes \cite{foster2019orthogonal}, and Besov, Sobolev, Hölder-type function classes \cite{nickl2007bracketing}. $L_2$ consistency of the pseudo-outcome bias follows if $\widehat{q}$ and one of $\widehat{e}$ or $\widehat{\rho}$ are $L_2$ consistent. 

\textbf{Second-stage sharp rates.} If $\|\mathcal{E}^+_{\tau} (x)\|= o_p\left( n^{-1/(2+r)} \right)$ and the conditions of \cref{thm:erm} hold, the pseudo-outcome bias has a negligible contribution to the CATE bounds estimation error. Thus, the estimation error is equivalent to the error as if the nuisances were known, a result known as the ``quasi-oracle property" (\cite{nie2021quasi}). 
Because the pseudo-outcome bias involves the product of rates, it will be sufficient to ask all pseudo-outcome nuisances to be consistent at an $o_p(n^{-1/4})$ rate. We give an example of sufficient conditions for our estimator to be oracle efficient (the property we synonymously call ``quasi-oracle” in our main text) in \cref{sec:MoreERMResults}. 

\begin{figure*}[ht!]
     \centering
     \begin{subfigure}[b]{0.415\textwidth}
         \centering
         \includegraphics[width=\textwidth]{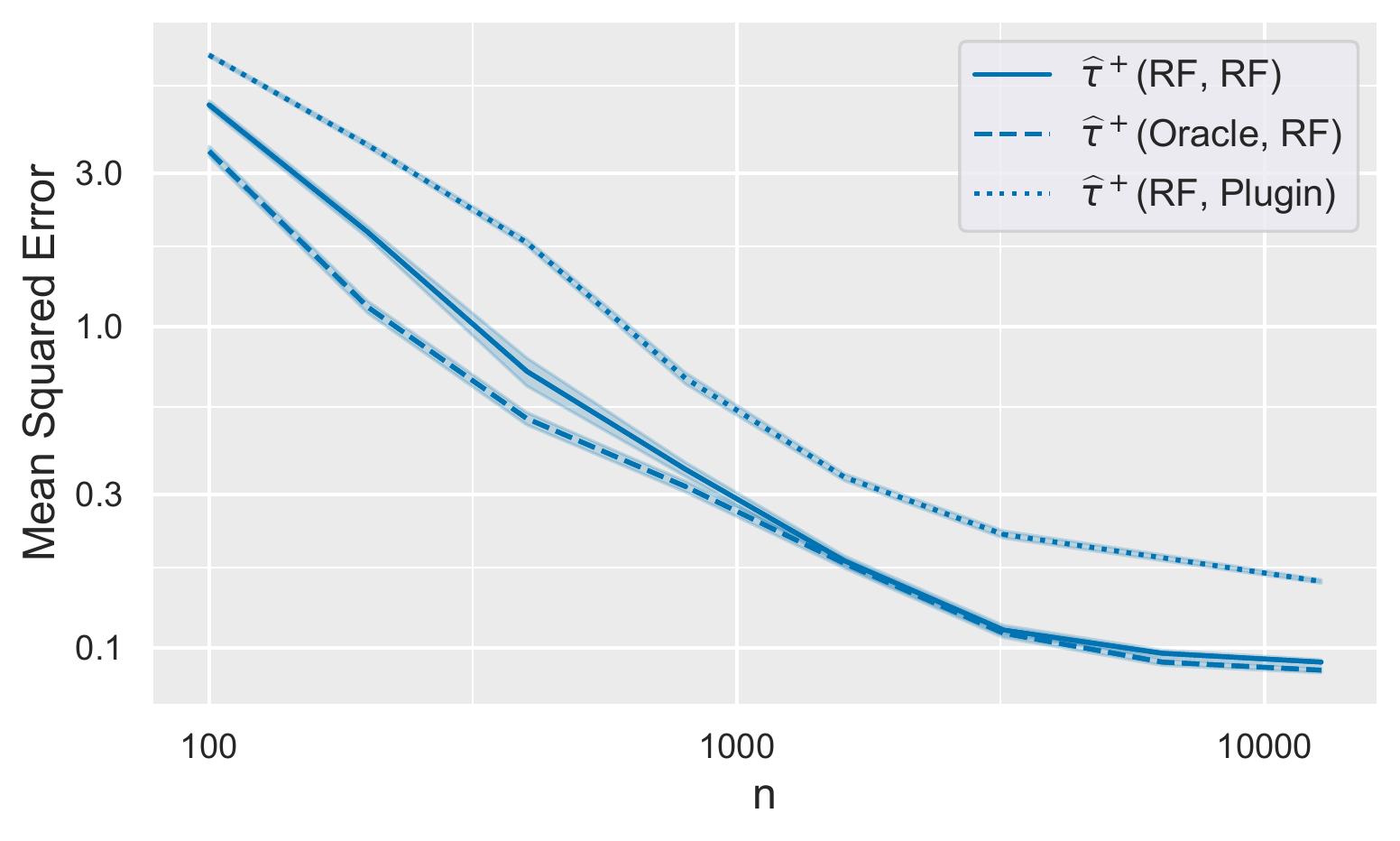}
         \caption{B-Learner and its oracle/plugin variants}
         \label{fig:rates}
     \end{subfigure}
     \begin{subfigure}[b]{0.56\textwidth}
         \centering
         \includegraphics[width=\textwidth]{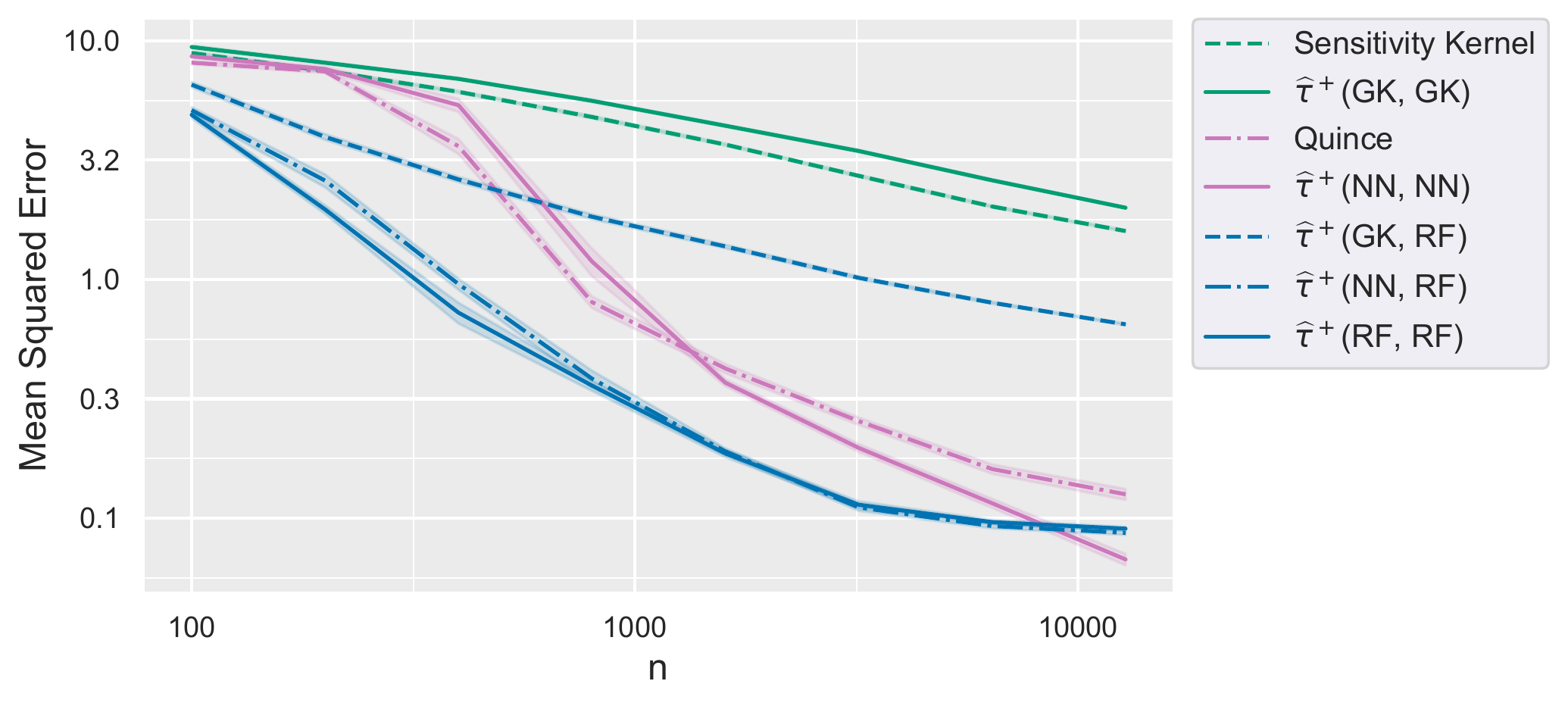}
         \caption{B-Learner, \textit{Sensitivity Kernel}, and \textit{Quince}}
         \label{fig:benchmarks}
     \end{subfigure}
     \vspace{-0.5em}
        \caption{Mean squared error (MSE) for different $\widehat{\tau}^{+}$ learners. Shaded regions depict plus/minus one standard error over 50 simulations.}
        \label{fig:cate-bound-sims}
     \vspace{-0.5em}
\end{figure*}

\textbf{Second-stage validity.} When the quantile estimates are inconsistent, we cannot apply \cref{thm:erm} directly. Still, we will have two chances to derive CATE bound estimates that are valid on average. In \cref{sec:LinearSmoothers}, we show that linear smoothers can yield stronger pointwise validity guarantees.
\begin{corollary}[ERM Validity on Average] \label{cor:ERMValidity} Assume the conditions of \cref{thm:erm} are satisfied and for all $f \in \mathcal{F}$ and $c \in \RR$ we have $f + c \in \mathcal{F}$. If $\|\widehat{q}_{+}(\cdot, 1) - \bar{q}_{+}(\cdot, 1)\| = o_P(1)$ and $\|\widehat{q}_{-}(\cdot, 0) - \bar{q}_{-}(\cdot, 0)\|=o_P(1)$ for a (potentially inconsistent) putative quantile function $\bar{q}$ and either $\|\widehat{e}-e^*\|=o_P(1)$ or both $\|\widehat{\rho}_{+}(\cdot, 1)-\rho_{+}^*(\cdot, 1, \bar{q}_+)\|=o_p(1)$ and $\|\widehat{\rho}_{-}(\cdot, 0)-\rho_{-}^*(\cdot, 0, \bar{q}_-)\|=o_p(1)$, then the estimated CATE bounds are valid on average in the sense that $\frac{1}{n} \sum_{i=1}^n \widehat{\tau}^{+}(X_i) - \tau^{+}(X_i) \geq -o_P(1)$.
\end{corollary}

\section{Experiments}\label{sec:exp-results}
In this section, we demonstrate our method on synthetic and semi-synthetic datasets, as well as on a real-world case study. We first benchmark the B-Learner using a synthetic example similar to that in \citet{kallus2019intervalEstimation}. We then illustrate how CATE bound estimators can be used for treatment deferral by using the hidden confounding variant of the IHDP dataset introduced by \citet{jesson2021quantifying}. For both sets of experiments, we compare with state-of-the-art methods proposed by \citet{kallus2019intervalEstimation} (\textit{Sensitivity Kernel}) and \citet{jesson2021quantifying} (\textit{Quince}\footnote{\citet{jesson2021quantifying} train an ensemble of several models, which is a computationally intensive task. For the purposes of this section, we do not ensemble any of the compared methods.}). We illustrate the usage of the B-Learner with real data through a case study of 401(k) eligibility effects on wealth. While we have focused our discussion on CATE upper bounds, our real data experiments also require estimating the CATE lower bounds we discuss in \cref{sec:lbResults}. Details about the data generation processes, specific model implementation, hyperparameter selection and validation procedures used are given in \cref{sec:add-exp-results}. We provide replication code at \url{https://github.com/CausalML/BLearner}. 

While the \textit{Sensitivity Kernel} approach uses Gaussian kernels and the \textit{Quince} model uses Bayesian neural networks, the B-Learner (\cref{alg:sharp-cate}) is flexible in the types of estimators allowed for both the first- and second-stage learners. We therefore compare three classes of nuisance and second-stage estimators: Random Forests (RF), Gaussian Kernels (GK), and Bayesian Neural Networks (NN). 
Whenever possible, we use the same hyperparameters and validation routine across models. For example, the B-Learner with NN nuisances uses the exact same neural networks as \textit{Quince}.

We denote the upper bound given by the B-Learner output (\cref{alg:sharp-cate}) by $\widehat{\tau}^{+}(\text{\{1\textsuperscript{st} stage\}}, \text{\{2\textsuperscript{nd} stage\}})$ (e.g. $\widehat{\tau}^{+}(RF, RF)$) to indicate the type of first- and second-stage learners used. For insight into the theoretical properties of our estimator, we also provide an oracle first-stage estimator $\widehat{\tau}^{+}(\text{Oracle}, \text{\{2\textsuperscript{nd} stage\}})$ which uses the true nuisances in the pseudo-outcome calculation, as well as a ``plug-in'' estimator $\widehat{\tau}^{+}(\text{\{1\textsuperscript{st} stage\}}, \text{Plugin})$ which plugs in the estimated nuisances into the expressions from \cref{result:sharp-cate-bounds}.

\subsection{Simulated Data}\label{sec:synthetic}

Our synthetic dataset is sampled as follows:
\begin{align*}
    &X \sim \text{Unif}([-2, 2]^{5}),\quad
    A \mid X  \sim \text{Bern}(\sigma(0.75X_0+0.5)),\\
    &\ts Y \sim \mathcal{N}((2A-1)(X_0+1)-2\sin\left((4A-2)X_0, 1\right),
\end{align*}
where $\sigma$ is the sigmoid function. We wish to provide an estimate $\widehat{\tau}^{+}(x)$ for the CATE upper bound under a level of confounding given by $\log \Lambda = 1$. With this simulation, it is straightforward to obtain the true nuisances $e^*, \mu^*, \rho^*$. These, along with \cref{result:sharp-cate-bounds}, allow us to determine the true value $\tau^{+}(x)$ of the upper bound.
We run $50$ simulations for sample sizes $n=100, 200, 400, ...,12800$ and evaluate the different models on a fixed test set of $400$ data points initially drawn at random. We compare the mean squared error (MSE) performance of each estimator with respect to the true bound and depict our findings in \cref{fig:cate-bound-sims}. 

In \cref{fig:rates}, we study the MSE convergence rates of the $\widehat{\tau}^{+}(RF, RF)$ estimator, along with its oracle and plug-in variants. The convergence rate of our estimator matches the rate of the oracle estimator. That is, \cref{alg:sharp-cate} with more than a few hundred observations performs essentially as well as if the estimator had access to the true, oracle nuisances. This confirms our theoretical results from \cref{thm:erm} in that small errors in the nuisance estimation lead to second-order errors in $\widehat{\tau}(x)$. Moreover, we see that the simple plug-in estimator suffers from so-called plug-in bias for every value of $n$, as anticipated. The B-Learner MSE improvement slows for large $n$, which we expect reflects our use of rules-of-thumb to extrapolate hyperparameters to large samples.

\begin{figure}[t]
    \centering
    \includegraphics[width=0.9\linewidth]{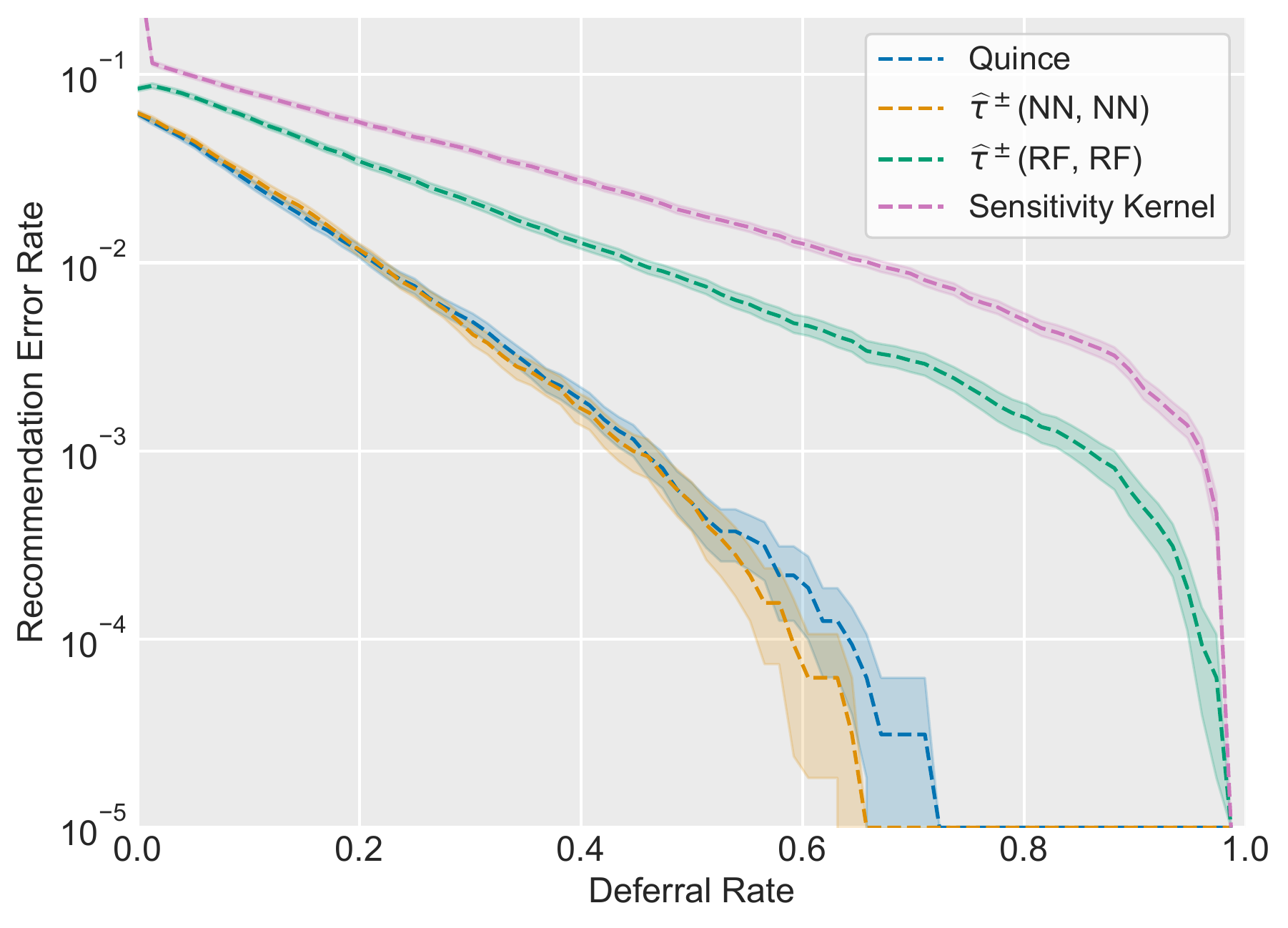}
    \vspace{-0.5em}
    \caption{IHDP Hidden Confounding: Error recommendation rate for different values of the percentage of deferred points. The x-axis represents different levels of practitioner caution by varying the percentage of recommendations deferred.}
    \vspace{-1em}
    \label{fig:ihdp}
\end{figure}

In \cref{fig:benchmarks}, we benchmark our estimator against \textit{Sensitivity Kernel} and \textit{Quince} for various first- and second- stage combinations. We see that using the same nuisances (GKs and NNs, respectively) leads to our method performing comparably with competitors. However, the B-Learner with NN or GK first stages and with RF second stage learners performs better than the state-of-the-art methods. This result underscores the importance of flexibility in choosing nuisance estimators, a key property of our method.
 
\begin{figure}[t]
  \begin{subfigure}[b]{0.43\columnwidth}
    \includegraphics[width=\linewidth]{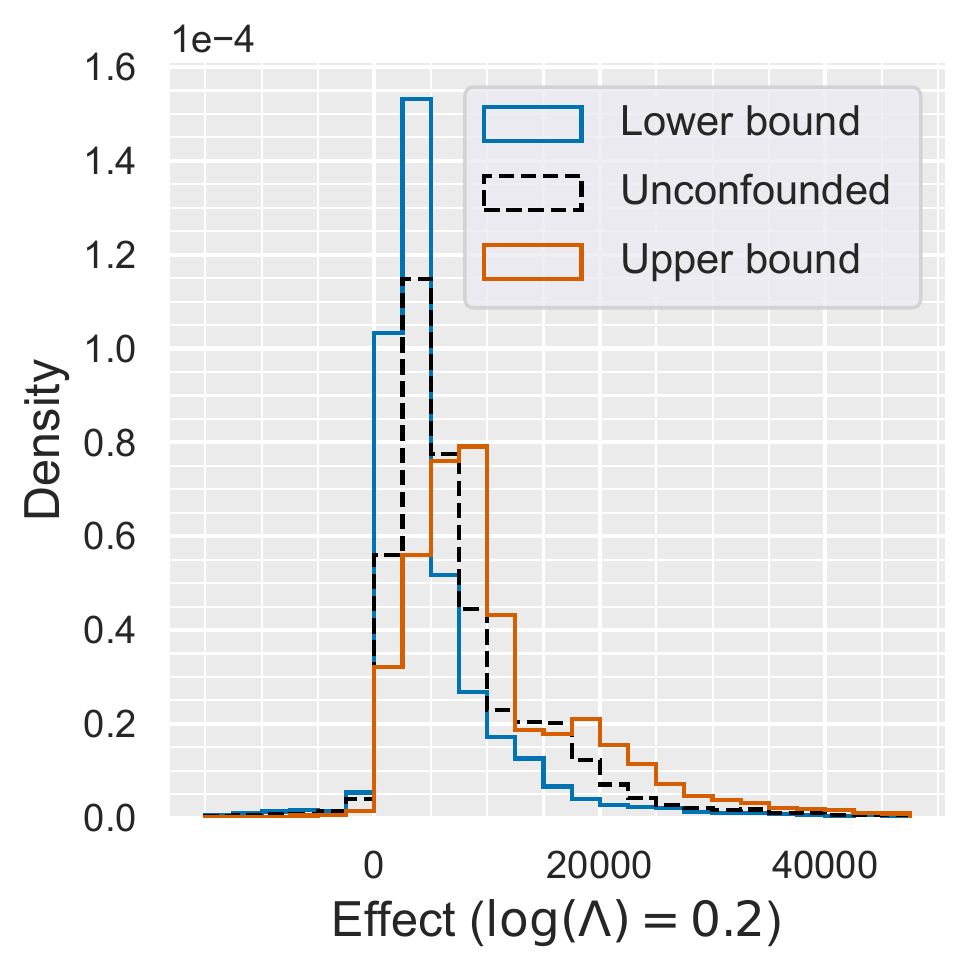}
    \caption{}
    \label{fig:401k_results_a}
  \end{subfigure}
  \begin{subfigure}[b]{0.52\columnwidth}
    \includegraphics[width=\linewidth]{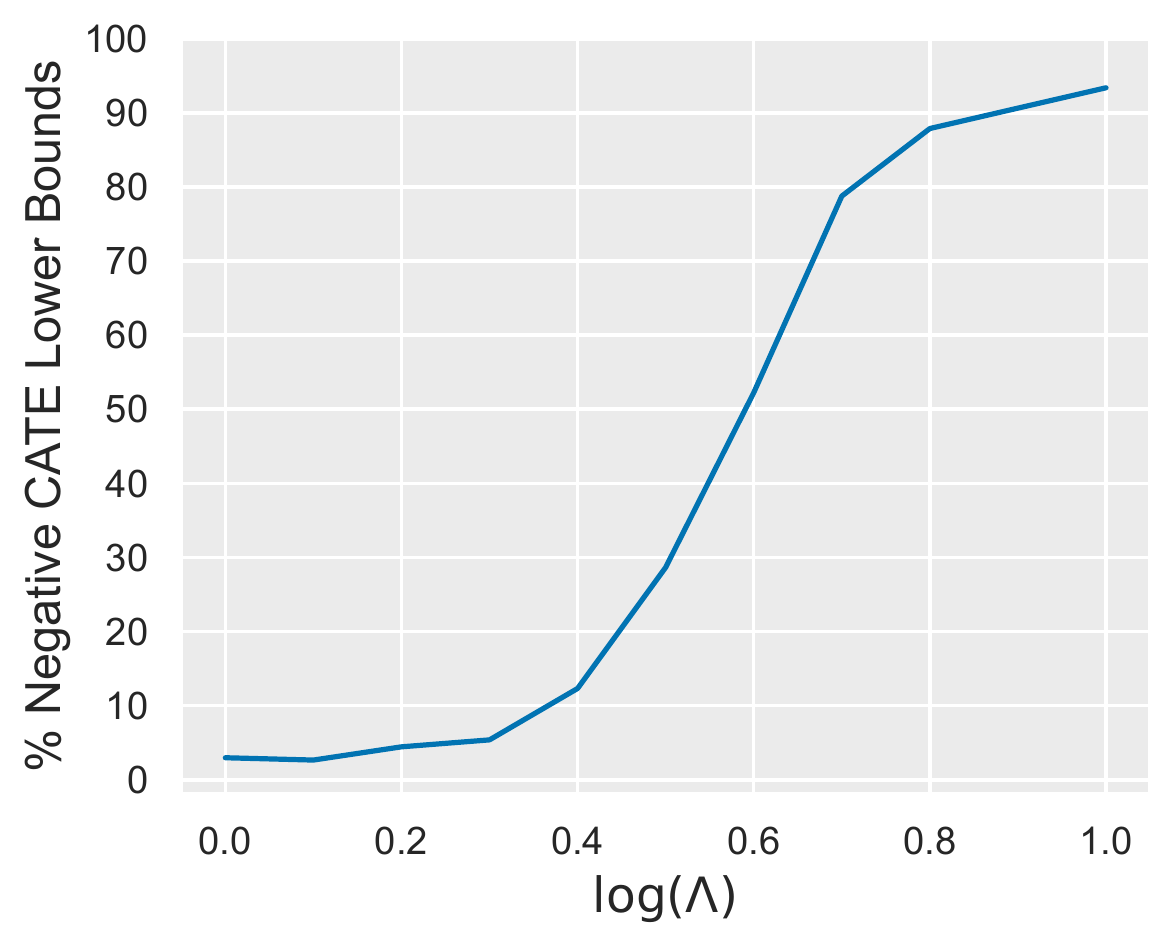}
    \caption{}
    \label{fig:401k_results_b}
  \end{subfigure}
  \caption{Bounds on the effect of 401(k) eligibility on financial wealth. (\ref{fig:401k_results_a}): Example of lower and upper bound effect distributions for $\log\Lambda = 0.2$. (\ref{fig:401k_results_b}): The percentage of lower bounds for which the effect is negative as a function of $\log(\Lambda)=0.1,...,1.0$. }
  \label{fig:401k_results}
  \vspace{-1em}
\end{figure}

\subsection{IHDP Hidden Confounding}

We now show how the B-Learner can be used for other causal inference tasks, such as informing \emph{deferral policies} for treatment recommendations. We replicate the experiment from \citet{jesson2021quantifying} on IHDP Hidden Confounding. The dataset is multi-dimensional, has low overlap, and has hidden confounding due to a single covariate being hidden from the training models. The dataset contains synthetic potential outcomes generated according to the response surface B described by \citet{hill2011bayesian}. We use the same deferral policy as in \citet{jesson2021quantifying}, namely, the policy simulates either recommending treatment or  deferral to an expert. We make a treatment recommendation (either $A=0$ or $A=1$, according to the sign of CATE estimate) if and only if the predicted CATE interval excludes zero. 

We measure model performance in terms of recommendation error rate across multiple deferral rates. The deferral rate is the fraction of observations for which we defer the action decision to the expert. The error rate is the percentage of  observations for which we recommend the wrong treatment, among those in which we did not defer.
Note that in this experiment we know the best treatment for each unit since we simulate both potential 
outcomes, although these effects do not correspond to a sharp bound under Assumption \ref{assumption:msm}. 

We compare two different variants of our B-Learner: $\widehat{\tau}^{\pm}(RF, RF)$ and $\widehat{\tau}^{\pm}(NN, NN)$ with \textit{Sensitivity Kernel} and \textit{Quince}. 
We see in \cref{fig:ihdp} that the RF B-Learner outperforms the GK-based \textit{Sensitivity Kernel} method, and that the best performing methods are the NN B-Learner and \textit{Quince} which perform very similarly. 

\subsection{Impact of 401(k) Eligibility on Wealth Distribution}\label{sec:real}

We apply the B-Learner to illustrate the impact of hidden confounding in a study of 401(k) eligibility and its effects on financial wealth. We use the real-world dataset from \citet{chernozhukov2004effects} that draws on the 1991 Survey of Income and Program Participation.  The treatment of interest is 401(k) eligibility, while the target outcome is the net financial assets of an individual (taken as the aggregate of 401(k) balance, bank accounts and interest-earning assets minus non-mortgage debt). 

This 401(k) eligibility dataset has been used in many analyses \cite{poterba1994401,chernozhukov2004effects}, often assuming unconfoundedness holds given observed covariates and finding a strong positive effect. 
However, unconfoundedness is an untestable assumption, so here we explore the uncertainty in the (conditional) treatment effects under varying degrees of hidden confounding. To that end, we apply the B-Learner algorithm repeatedly for different settings of $\Lambda$: $\log\Lambda=0.1,0.2,...,1.0$. The nuisances are all estimated using Random Forest models with hyperparameters as in \citep{chernozhukov2018double}. We also estimate the CATE under assumed unconfoundedness ($\log\Lambda=0$ which corresponds to the DR-Learner \cite{kennedy2020optimal}). 

In \cref{fig:401k_results}, we plot the distribution of predicted conditional effects on the 9,915 observations for $\log\Lambda = 0.2$ as well as the fraction of negative lower bound effects (frequency of $\mathbb{I}(\widehat{\tau}^-(x) \leq 0)$) as we vary $\Lambda$. For lower values of $\Lambda$, the majority of lower bounds are still positive, which means that under those levels of confounding, most true conditional treatment effects are still positive. However, as we increase $\Lambda$, more and more of the true effects could be negative as the lower bound is comprised of mostly negative effects. 
For example, at $\log(\Lambda)=0.6$, about half of the CATE lower bounds are negative which is to be interpreted as: if the data were truly confounded at this level, 50\% of the effects measured as positive could in reality have been negative due to unobserved confounders.
Regardless of what $\Lambda$ level is most appropriate here, we see the B-Learner is a powerful tool for practitioners who wish to conduct what-if experiments for potential unobserved confounding.  

\section{Conclusion}

We presented the B-Learner, a meta-learner for estimating bounds on the CATE function. The B-Learner can use any learning method as its base learners, including random forests and neural nets. We showed that the B-Learner provides bound estimates that are valid, sharp, robust, and have quasi-oracle rate properties, making it (to the best of out knowledge) the first CATE sensitivity analysis method with all these properties. Experiments validate our theoretical findings, show that the B-learner is comparable in performance to existing state-of-the-art methods, and demonstrate it can be used with real-world data to gain insight into the uncertainty of estimated causal effects. 

\section*{Acknowledgements}
 This material is based upon work supported by the National Science Foundation under Grant No. 1846210. M.O. was supported by U.S. Department of Energy, Office of Science, Office of Advanced Scientific Computing Research, under Award Number DE-SC0023112. M.G and U.S. were supported by Israeli Science Foundation, grant number 1950/19. Some of this material is based upon work by J.D. supported by the National Science Foundation Graduate Research Fellowship Program under Grant No. DGE-2039656. The authors would like to thank the anonymous reviewers and Rebecca Dorn for useful discussion and feedback.

\bibliography{references}
\bibliographystyle{icml2023}

\newpage
\appendix
\onecolumn


\textbf{Note: } Throughout the appendix, we use $\pm$ notation to encode either upper/ lower bounds results. This allows us to unify upper/ lower results and proofs at the cost of some readability. 

\section{Notation}\label{sec:notation}

We summarize the notation we use throughout this work in \cref{tab:notation}. In addition, note that we use upper case letters (e.g. $X$) to denote random variables and lower case letters (e.g. $x$) to refer to specific values of a random variable.

\begin{center}
\captionof{table}{Notation}
\vspace{0.5em}
\begin{tabular}{l | l } 
$X$ & The observed covariates in $\mathbb{R}^d$\\
$A$ & A binary treatment ($A \in \{0, 1\}$)\\
$Y$ & The outcome\\
$Z$ & $(X,A,Y)$ which is drawn from an observed distribution $\Po$\\
$Y(1)$, $Y(0)$ & Real-valued treated and untreated potential outcomes, respectively\\
$U$ & The unobserved confounder in $\mathbb{R}^k$\\
$\Pf$ & An unobservable distribution  over $(X, A, Y(1), Y(0), U)$\\
$\alpha$ &  $\frac{\Lambda}{\Lambda+1}\in[0.5, 1)$ for $\Lambda \geq 1$\\
$\{b\}_+$, $\{b\}_-$ &$\max\{b, 0\}$ ,$\min\{b, 0\}$ respectively, for a real number, $b$  \\
$b\lesssim d$ & $b\leq Cd$, for $b, d\in \RR$, and for some universal constant $C$ \\
$g^*$ & The true value of a function $g$\\
$\bar{g}$ & A putative value of a function $g$\\
$\widehat{g}$ & An estimated value of a function $g$ from data\\
$\|g\|:=\EE_F[g(z)^2]^{1/2}$ & The $L_2$ norm of $g$ given a probability distribution $F(z)$ and a function $g(z)$ \\
$+, -$ & Indicators to symbolize upper and lower bounds, respectively\\
$\pm$, $\mp$ & Symbols signal that an equation should be read twice,\\
& once with $\pm=+, \mp=-$ and once with $\pm=-, \mp=+$\\
& E.g.: $a^{\pm}=b^{\pm}+c^{\mp}$ encodes two equalities: $a^+=b^++c^-$ and $a^-=b^-+c^+$\\
$e(x)$ & The observed propensity score $P(A = 1 \:|\: X=x)$ \\
$e(x, u)$ & The full propensity score $\Pf(A = 1  \:|\: X=x, U=u)$\\ 
$F(y \:|\: x, a)$ & The conditional outcome distribution, $P(Y\leq y  \:|\: X=x, A=a)$\\
$f(y\mid x, a)$ & The conditional outcome density, $\frac{d}{d y}F(y\mid x, a)$\\
$\mu^*(x, a)$ & $\EE[Y\mid X=x, A=a]$, outcome regression\\ 
$ q_{c}^*(x, a)$ & $\inf\{\beta: F(\beta\mid x, a)\geq c\}$, conditional outcome quantile\\
$q_+^*(x, a)$ & $ q_{\alpha}^*(x, a)$, shorthand $\alpha^{\text{th}}$ quantile notation\\
$q_-^*(x, a)$ & $q_{1-\alpha}^*(x, a)$, shorthand $(1-\alpha)^{\text{th}}$ quantile notation \\
$H_{\pm}(z, \bar{q})$ & $\bar{q}(x, a) + \frac{1}{1-\alpha} \left\{y-\bar{q}(x, a)\right\}_{\pm}$, Conditional Value at Risk pseudo-outcome \\
$\text{CVaR}_{\pm}(x, a)$ & $\EE[ H_{\pm}(z, q_\pm^*)\mid X=x, A=a]$, the Conditional Value at Risk\\
$\text{CVaR}_+(x, a)$ & The expectation above the $(1-\alpha)$ quantile\\
$\text{CVaR}_-(x,a)$ & The expectation below the $\alpha$ quantile\\
$R_\pm(z, \bar{q})$ & $\Lambda^{-1}y+(1-\Lambda^{-1}) H_{\pm}(z, \bar{q})$, pseudo-outcome for the (conditional) unobserved potential outcome \\
$\rho_{\pm}^*(x, a, \bar{q})$ & $\EE[ R_{\pm}(z, \bar{q})\mid X=x, A=a]$, the (conditional) expected unobserved potential outcome\\
$\rho_\pm^*(x, a)$ & A shorthand for $\rho_{\pm}^*(x, a, q_\pm)^*$, the $\rho_{\pm}^*$ function evaluated at the true conditional quantiles $q_{\pm}^*$ \\
\hline
\multicolumn{2}{l}{CATE Bounds Pseudo-Outcomes} \\
\hline
$\phi^{+}_1(Z, \widehat{\eta})$ & $AY + (1-A)\widehat{\rho}_{+}(X, 1)\ts+\frac{\left(1-\widehat{e}(X)\right)A}{\widehat{e}(X)}\cdot \left(R_{+}(Z, \widehat{q}_{+}(X, 1))-\widehat{\rho}_{+}(X, 1)\right)$\\
$\phi^{-}_0(Z, \widehat{\eta})$ & $(1-A)Y + A\widehat{\rho}_{-}(X, 0)+ \ts\frac{\widehat{e}(X)(1-A)}{\left(1-\widehat{e}(X)\right)}\cdot \left(R_{-}(Z, \widehat{q}_{-}(X, 0))-\widehat{\rho}_{-}(X, 0)\right)$\\
$\phi^{+}_{\tau}(Z, \widehat{\eta})$ &  $\phi^{+}_1(Z, \widehat{\eta}) - \phi^{-}_0(Z, \widehat{\eta})$\\
\hline
\end{tabular}
\label{tab:notation}
\end{center}

\section{Results for CATE Lower Bounds}\label{sec:lbResults}

The results for the CATE lower bound $\tau^-(x)$ can be obtained by interchanging $+$ and $-$ symbols in the nuisances and/or replacing $A$ with $1-A$. We state them here for completeness.

\textbf{CATE lower bounds identification (Result 1).} The sharp CATE lower bound is given by $\tau^-(x)=Y^-(x, 1)-Y^+(x, 0)$, where the relevant bounds on the conditional average potential outcomes can be expressed as:
\begin{align*}
    Y^-(x, 1) & = e^*(x)\mu^*(x,1)+(1-e^*(x))\rho^*_-(x, 1),\\
    Y^+(x, 0) & = (1-e^*(x))\mu^*(x,0)+e^*(x)\rho^*_+(x, 0).
\end{align*}
Thus, the lower bounds can be expressed as a convex combination of quantities that can be estimated from the observed data alone, i.e. they are \textit{identifiable} from data. 

\textbf{Pseudo-outcomes for CATE lower bounds (Definition 2).} Let $\widehat{\eta} = (\widehat{e}, \widehat{q}_{+}(\cdot, 0), \widehat{q}_{-}(\cdot, 1), \widehat{\rho}_{+}(\cdot, 0), \widehat{\rho}_{-}(\cdot, 1))\in \Xi$ be a set of nuisances. The pseudo-outcomes for the bounds $Y^-(x, 1), Y^+(x, 0)$ and $\tau^{-}(x)$ are given by: 
\begin{align*}
        & \phi^{-}_1(Z, \widehat{\eta}) =  AY + (1-A)\widehat{\rho}_{-}(X, 1)  +\frac{\left(1-\widehat{e}(X)\right)A}{\widehat{e}(X)}\cdot \left(R_{-}(Z, \widehat{q}_{-}(X, 1))-\widehat{\rho}_{-}(X, 1)\right),\\
         & \phi^{+}_0(Z, \widehat{\eta}) =  (1-A)Y + A\widehat{\rho}_{+}(X, 0) + \frac{\widehat{e}(X)(1-A)}{\left(1-\widehat{e}(X)\right)}\cdot \left(R_{+}(Z, \widehat{q}_{+}(X, 0))-\widehat{\rho}_{+}(X, 0)\right),
    \\
    &\phi^{-}_{\tau}(Z, \widehat{\eta}) =  
     \phi^{-}_1(Z, \widehat{\eta}) - \phi^{+}_0(Z, \widehat{\eta}).
\end{align*}

\textbf{Validity and sharpness for CATE lower bounds.} We call lower bound estimates $\widehat{\tau}^-(x)$ \textit{valid} if $\widehat{\tau}^-(x) - \tau^-(x)\leq -o_P(1)$. Similarly, the lower bound estimates $\widehat{\tau}^-(x)$ are \textit{sharp} if $\widehat{\tau}^-(x)=\tau^-(x)+o_P(1)$. 

\textbf{Pseudo-outcome bias for CATE lower bounds.} The absolute bias of the CATE lower bound pseudo-outcome has the form:
\begin{align*}
\begin{split}
    \left| \mathcal{E}^{-}_\tau(x; \widehat{\eta}) \right|  &\lesssim |\widehat{e}(x)-e^*(x)|\;|\widehat{\rho}_{-}(x, 1)-\rho_{-}^*(x, 1)| \\ 
    & \quad + |\widehat{e}(x)-e^*(x)|\;|\widehat{\rho}_{+}(x, 0)-\rho_{+}^*(x, 0)| \\
    & \quad + (\widehat{q}_{-}(x, 1) - q_{-}^*(x, 1))^2 \\
    & \quad + (\widehat{q}_{+}(x, 0) - q_{+}^*(x, 0))^2.
    \end{split} 
\end{align*}
whereas the signed bias bound is given by:
\begin{align*}
    \mathcal{E}^-_\tau(x; \widehat{\eta}) & \lesssim |\widehat{e}(x)-e^*(x)|\;|\widehat{\rho}_{-}(x, 1)-\rho_{-}^*(x, 1, \widehat{q}_{-})| \\ 
    & \quad - |\widehat{e}(x)-e^*(x)|\;|\widehat{\rho}_{+}(x, 0)-\rho_{+}^*(x, 0, \widehat{q}_{+})|. 
 \end{align*} 
\hfill \qedsymbol{}

The proofs of the theorems and corollaries in the paper (\cref{sec:proofs}) are unified across lower/upper bounds by using the $\pm$ notation described above. For example, we will write the consolidated pseudo-outcome bias bounds as:
\begin{align*}
    \left| \mathcal{E}^\pm_a(x; \widehat{\eta}) \right|  &\lesssim |\widehat{e}(x)-e^*(x)|\;|\widehat{\rho}_{\pm}(x, a)-\rho^*_{\pm}(x, a)| \\
    & \quad + (\widehat{q}_{\pm}(x, a) - q^*_{\pm}(x, a))^2\\
    \mp \mathcal{E}^\pm_a(x; \widehat{\eta}) & \lesssim |\widehat{e}(x)-e^*(x)|\;|\widehat{\rho}_{\pm}(x, a)-\rho^*_{\pm}(x, a, \widehat{q}_{\pm})| . 
\end{align*}
which, together with $\mathcal{E}^\pm_\tau(x;\widehat{\eta})=\mathcal{E}^\pm_1(x;\widehat{\eta})-\mathcal{E}^\mp_0(x;\widehat{\eta})$, yield the bias bounds for the lower and upper CATE bounds.

\section{More Estimation Results}

\subsection{More ERM  Results}\label{sec:MoreERMResults}

\begin{corollary}[Conditions for ERM Oracle Efficiency]\label{ex:erm}
    Let $\Fcal$ be a class of $\beta$-smooth functions in $d$ dimensions (i.e. H\"older) and let $e, \rho_\pm, q_\pm$ be $\gamma_e$, $\gamma_\rho$, and $\gamma_q$-smooth functions, respectively. Then, the $L_2$ error rate of \cref{alg:sharp-cate} is $O_p\big(n^{-1/(2+d/\beta)} + n^{-2/(2+d/\gamma_q)} + n^{-(1/(2+d/\gamma_e)  +1/(2+d/\gamma_\rho)}\big)$. Furthermore, if $\gamma_q\geq \frac{d/2}{1+d/\beta}$ and $\gamma_\rho\gamma_e\geq \frac{d^2}{4}-\frac{(\gamma_\rho+d/2)(\gamma_e+d/2)}{1+2\beta/d}$, our estimator is oracle efficient in the sense that the leading order error is that of the oracle estimator, $\widehat{\EE}_n[\phi^{\pm}_{\tau}(Z, \eta^*)\mid X=x]$. 
\end{corollary}

\subsection{Doubly Robust-Style Smoothing Estimators}\label{sec:LinearSmoothers}

We now study the behavior of \cref{alg:sharp-cate} with a DR Learner-style smoothing estimator as the second-stage learner. This technique was introduced in \citet{kennedy2020optimal} and includes a wide range of estimators satisfying certain stability conditions, with linear smoothers as the archetype of this class. In this section, we analyze a generic linear smoother defined as follows:
\begin{equation*}\label{eq:linsmooth}
    \widehat{\EE}_n\left[\widehat{\phi}^{\pm}_{\tau}\mid X=x\right] = \frac{1}{n}\sum_{i=1}^n w_i(x) \widehat{\phi}^{\pm}_{\tau, i}
\end{equation*}
where the $w_i(x)$'s are weights learned on a different sample than $\widehat{\phi}^{\pm}_{\tau, i}$ (which can be achieved by sample splitting). Under mild regularity assumptions, this estimator can yield stronger guarantees in the form of pointwise error bounds.

\begin{theorem}[Rates for Linear Smoothing Estimators]\label{thm:drsmooth} Assume the conditions of \cref{asm:bounded}.
Then:
\begin{align*}
    \left|\widehat{\tau}^\pm(x) - \tau^\pm(x)\right| & \lesssim \left|\widetilde{\tau}^\pm(x) - \tau^\pm(x)\right| + b_n^\pm(x) + O_p\left( \left( \| \widehat{\phi}_\tau^{\pm} - \phi_\tau^{\pm} \|_{w^2} + o_p(1) \right) \left( \frac{1}{n^2} \sum_{i=1}^n w_i(x)^2 \right)^{1/2} \right) 
\end{align*}
where $\widetilde{\tau}^\pm(x)$ corresponds to the linear smoother procedure with oracle first-stage nuisances, $\| \cdot \|_{w^2}$ is the empirical $w_i(x)^2$-weighted distance of \citet{kennedy2022towardsOptimal}, 
and the $b_n^\pm(x)$ bias function is of the form:
\begin{align*}
    b_n^\pm(x) & = \left| \frac{1}{n} \sum_{i=1}^n w_i(x) \mathcal{E}_\tau^{\pm}(X_i; \widehat{\eta}) \right|
\end{align*} 
\end{theorem}

\textbf{Second-stage sharp consistency and robustness.} $\tilde{\tau}$ consistency follows under weak conditions like $\frac{1}{n} \sum_{i=1}^n |w_i(X_i)|\leq C$ \cite{stone1977consistent}. Thus, we can state corollaries that prove consistent estimation of sharp bounds under either strong restrictions on weights or strong requirements on consistency. We show one such corollary for a wide class of linear smoothers that includes linear and ridge regression, local polynomial and RKHS regression, kernel estimators, and some tree methods \cite{wasserman2006all}. In this corollary, we ask for uniform nuisance consistency to make the bias term $b_n^{\pm}$ tend to zero. 

\begin{corollary}[Pointwise Consistency of Sharp CATE Bounds] \label{cor:PointwiseLinearSmoother} Assume the conditions of \cref{thm:drsmooth} are satisfied, the $\frac{w_i(x)}{n}$ weighting functions satisfy the requirements of \citet{stone1977consistent} Theorem 1, and $\frac{1}{n} \sum |w_i(x)| = O_p(1)$. If $\widehat{q}_{\pm}$ and either $\widehat{e}$ or $\widehat{\rho}$ are uniformly consistent, then $\widehat{\tau}^\pm(x)$ converges to the true pointwise sharp CATE bounds. 
\end{corollary}

\textbf{Second-stage sharp rates.} Take $\tau^\pm, e, \rho_\pm, q_\pm$ be $\gamma_q$  to be H\"older with smoothness $\beta$, $\gamma_e$, $\gamma_\rho$, and $\gamma_q$. Then, the pointwise error rate of \cref{alg:sharp-cate} is $O_p\big(n^{-1/(2+d/\beta)} + n^{-2/(2+d/\gamma_q)} + n^{-(1/(2+d/\gamma_e)  +1/(2+d/\gamma_\rho)}\big)$ and the estimator will be oracle efficient, though the error bounds here are \textit{pointwise} (local), whereas the ERM-based bounds are $L_2$ (global).

\textbf{Second-stage validity.} The linear smoothers also have pointwise validity. Unlike in the ERM case where the best model fit to conservative bounds might extrapolate to invalid bounds for some regions of the covariates, the linear smoothers will have pointwise validity guarantees.  
\begin{corollary}[Pointwise Validity of Lax CATE Bounds] \label{cor:LinearSmootherPointwiseValidity}
Assume the conditions of \cref{thm:drsmooth} are satisfied, $w_i(x)$ satisfies the requirements of Theorem 1 in \citet{stone1977consistent}, and $\frac{1}{n} \sum_{i=1}^n |w_i(x)| = O_p(1)$. If $\widehat{q}_{\pm}$ is uniformly consistent to some limiting quantile $\bar{q}_{\pm}$ and $\widehat{e}$ is uniformly consistent for $e^*$ or $\widehat{\rho}_\pm$ is uniformly consistent for $\rho_\pm^*(X, A, \bar{q}_{\pm})$ and $f(\bar{q}_\pm(x, a) \mid x, a) > 0$. Then the estimated bounds are pointwise valid in the sense that $\pm \left( \widehat{\tau}^\pm(x) - \tau^\pm(x) \right) \geq -o_P(1)$.
\end{corollary}

\section{Proofs}\label{sec:proofs}
Note: we assume throughout that $X$, $U$, $Y(0)$, and $Y(1)$ to have probability measures absolutely continuous w.r.t. the Lebesgue measure so that we can condition on the event $X=x$.
\begin{proof}[Proof of \cref{thm:condneym}]\label{proof:condneym} We start with the bound for the unsigned bias.
Consider the $Y^{+}(X, 1)$ bound for simplicity. We first show that our problem fits into the framework of \citet{kallus2022robust} since the estimand and the oracle nuisances are the solutions of following conditional moment restrictions:
\begin{align*}
    & \EE[AY + (1-A)\rho_{+}^*(X,1)-Y^{+}(X, 1)\mid X] = 0 \tag{Estimand moment}\\
    & \EE[R_+(Z, q_+^*(X, 1))-\rho_{+}^*(X, 1)\mid X, A=1] = 0 \tag{Modified outcome moment}\\
    & \EE[\alpha - \II(Y\leq q_+^*(X, 1))\mid X, A=1] = 0 \tag{Quantile moment}
\end{align*}
Let $\nu_1$ be the nuisance set corresponding to this set of moments (as defined in \citet{kallus2022robust}). Then $\nu_1^*(X)=(Y^{+}(X, 1), \rho_{+}^*(X, 1), q_{+}^*(X, 1))$. This is different from $\eta^*$ since the propensity does not have an estimating conditional moment. The Jacobian of the moments with respect to $\nu_1^*$ is thus given by:
\begin{align*}
    J^*_1(X) = \begin{pmatrix}
    -1 & 1-e^*(X) & 0\\
    0 & -1 & 0\\
    0 & 0 & -f(q_+^*(X, 1)\mid X, 1)
    \end{pmatrix}
\end{align*}
where $f(y\mid x, 1)$ be the conditional density at a point $y$ for $a=1$. The first row of the inverse is then given by $\alpha^*_1(X):=(J^*_1(X))_1^{-1}= (-1, e^*(X)-1, 0)$. Thus, using the pseudo-outcome in Definition 3 of \citet{kallus2022robust}, replacing $\nu_1^*, \alpha_1^*$ with their estimated counterparts $\widehat{\nu}_1(X)=(\widehat{Y}^{+}(X, 1), \widehat{\rho}_+(X, 1), \widehat{q}_+(X, 1))$, $\widehat{\alpha}_1(X)=(-1, \widehat{e}(X)-1, 0)$, and noting that the first moment is conditional only on $X$, we obtain the pseudo-outcome:
\begin{align*}
 \phi^{+}_1(Z, \widehat{\eta}) =  AY + (1-A)\widehat{\rho}_{+}(X, 1) +\frac{\left(1-\widehat{e}(X)\right)A}{\widehat{e}(X)}\cdot \left(R_+(Z, \widehat{q}_+)-\widehat{\rho}_{+}(X, 1)\right)
\end{align*}

as desired. Therefore, our \cref{asm:bounded} is a direct application of the boundedness assumption (Assumption 1) in \citet{kallus2022robust} and the bound for the unsigned bias follows largely from their Theorem 1. We first note that the results of Theorem 1 in \citet{kallus2022robust} also hold pointwise (see the proof in their Appendix A). It now remains to calculate the $H$ and $G$ matrices in their Assumption 1:
\begin{align*}
    G = 
    \begin{pmatrix}
     1 & 1 & 0\\
     0 & 1 & 0\\
     0 & 0 & 1
    \end{pmatrix}, \quad H = 
    \begin{pmatrix}
     0 & 0 & 0\\
     0 & 0 & 0\\
     0 & 0 & 1
    \end{pmatrix}
\end{align*}
since $G$ is just a binary mask for $J_1^*(X)$ and $H$ involves second order derivatives of the moments. Plugging these into the bound for the unsigned bias, we obtain:
\begin{align*}
    \left| \mathcal{E}^{+}_1(x; \widehat{\eta}) \right|  &\lesssim \sum_{i=1}^3 \sum_{j=1}^3 G_{ij}\left|\widehat{\alpha}_{1,i}(x)-\alpha_{1,i}^*(x)\right|\left|\widehat{\nu}_{1,j}(x)-\nu_{1,j}^*(x)\right| + \sum_{i=1}^3 \sum_{j=1}^3 H_{ij}\left|\widehat{\nu}_{1,i}(x)-\nu_{1,i}^*(x)\right|\left|\widehat{\nu}_{1,j}(x)-\nu_{1,j}^*(x)\right|\\
    &\lesssim |\widehat{e}(x)-e^*(x)|\;|\widehat{\rho}_{+}(x, 1)-\rho_{+}^*(x, 1)| + (\widehat{q}_+(x, 1) - q_{+}^*(x, 1))^2.
\end{align*}
The result for $\mathcal{E}^{+}_0(x;\widehat{\eta})$ follows from replacing $a=1$ with $a=0$ everywhere. The bound for $\mathcal{E}^-_a(x;\widehat{\eta})$ follows from writing the corresponding conditional moments for $Y^{-}(X, a)$.

We now study the bound for the signed bias. We first take the expectation of $\phi_1^{+}(Z, \widehat{\eta})$:
\begin{align*}
    \EE\left[ \phi_1^{+}(Z, \widehat{\eta}) \mid X \right] & = \EE\left[ A Y + \left( 1 - \frac{A}{\widehat{e}(X)} \right) \widehat{\rho}_+(X, 1) + A \frac{1-\widehat{e}(X)}{\widehat{e}(X)} R_+(Z, \widehat{q}_+) \mid X \right] \\
    & = e^*(X) \mu^*(X, 1) + \frac{\widehat{e}(X) - e^*(X)}{\widehat{e}(X)} \widehat{\rho}_+(X, 1) + \left( \frac{e^*(X)}{\widehat{e}(X)} - e^*(X) \right) \rho_+^*(X, 1, \widehat{q}_+) \\
    & = e^*(X) \mu^*(X, 1) + \left( 1 - \frac{e^*(X)}{\widehat{e}(X)} \right) \left( \widehat{\rho}_+(X, 1) - \rho_+^*(X, 1, \widehat{q}_+) \right) +  \left( 1 - e^*(X) \right)\rho_+^*(X, 1, \widehat{q}_+) 
\end{align*} 
As a result, we can write write:
\begin{align*}
    \EE\left[ \phi_1^{+}(Z, \widehat{\eta}) - \phi_1^{+}(Z, \eta^*) \mid X \right] & = \frac{\widehat{e}(X) - e^*(X)}{\widehat{e}(X)} \left( \widehat{\rho}_+(X, 1) - \rho_+^*(X, 1, \widehat{q}_+) \right) + \left( \rho_+^*(X, 1, \widehat{q}_+) - \rho_+^*(X, 1) \right) \left( 1 - e^*(X) \right) 
\end{align*} 

Recall the CVaR property that $\rho_+^*(X, 1) = \inf_{\bar{q}} \rho_{+}^*(X, 1, \bar{q})$, so that $\rho_+^*(X, 1, \widehat{q}_+) \geq \rho_+^*(X, 1)$.

Therefore we have:
\begin{align*}
    -\mathcal{E}_1^{+}(x; \widehat{\eta}) & = \EE\left[ \phi_1^{+}(Z, \eta^*) - \phi_1^{+}(Z, \widehat{\eta}) \mid X=x \right] \\
    & = -\frac{\widehat{e}(x) - e^*(x)}{\widehat{e}(x)} \left( \widehat{\rho}_+(x, 1) - \rho_+^*(x, 1, \widehat{q}_+) \right) - \left( \rho_+^*(x, 1, \widehat{q}_+) - \rho_+^*(x, 1) \right) \left( 1 - e^*(x) \right) \\
    & \leq -\frac{\widehat{e}(x) - e^*(x)}{\widehat{e}(x)} \left( \widehat{\rho}_+(x, 1) - \rho_+^*(x, 1, \widehat{q}_+) \right) \\
    & \lesssim - \left| \widehat{e}(x) - e^*(x) \right| \left|\widehat{\rho}_+(x, 1) - \rho_+^*(x, 1; \widehat{q}_+) \right|
\end{align*} 

The result for $Y^{+}(X, 0)$ follows by symmetry. The result for $Y^-(X, a)$ follows by negating $Y$, applying the argument, and negating the argument. Results for $\tau^{\pm}(X)$ follow by \cref{result:sharp-cate-bounds}. 
\end{proof}

\begin{proof}[Proof of \cref{thm:erm}]
Since we showed in the proof of \cref{thm:condneym} that our pseudo-outcomes fit into the framework of \citet{kallus2022robust} and our \cref{asm:bounded} maps to their Assumption 1, we can apply their Theorem 2 directly to our setting, yielding the statement of our theorem. 
\end{proof}

\begin{proof}[Proof of \cref{cor:ERMValidity}]
Using lax notation, choose an estimated $\widehat{f}^\pm \in \mathcal{F}$ to minimize Equation (\ref{eq:erm}) and then an estimated $\widehat{c}^\pm \in \RR$ such that $\widehat{f}^\pm + \widehat{c}^\pm$ is a minimizer of Equation (\ref{eq:erm}). By construction, we must have $\widehat{f}^\pm + 0$ is an optimizer.

If we differentiate (\ref{eq:erm}) with respect to $\widehat{c}^\pm$, evaluate it at $0$, and divide by $2$, we obtain the requirement on any optimizer that $\frac{1}{n} \sum_{i=1}^n \left( \widehat{\phi}^{\pm}_{\tau, i} - \widehat{f}^{\pm}(X_i) \right) = 0$. As a result:
\begin{align*}
    \pm \left( \frac{1}{n} \sum_{i=1}^n \widehat{\tau}^{\pm}(X_i) - \tau^{\pm}(X_i) \right) & = \pm \left( \frac{1}{n} \sum_{i=1}^n \widehat{\phi}^{\pm}_{\tau, i} - \tau^{\pm}(X_i) \right) 
\end{align*}

By applying Chebyshev's inequality to the average of zero-meaned bounded random variables $\widehat{\tau}^{\pm}(X_i) - \tau^{\pm}(X_i) - \mathcal{E}_\tau^{\pm}(X_i; \widehat{\eta})$, we can further obtain:
\begin{align*} 
    \pm \left( \frac{1}{n} \sum_{i=1}^n \widehat{\tau}^{\pm}(X_i) - \tau^{\pm}(X_i) \right) & = \pm \frac{1}{n} \sum_{i=1}^n \mathcal{E}_{\tau}^\pm(X_i; \widehat{\eta}) - O_p(n^{-1/2}) \\
    & \geq - \frac{1}{n} \sum O\left( | \widehat{e}(X) - e^*(X)| \sum_a | \widehat{\rho}_{\pm}(X, a) - \rho_{\pm}^*(X, a, \widehat{q}_{\pm}) | \right) - O_p(n^{-1/2}) \\
    & \geq - O\left( \| \widehat{e} - e^* \| \sum_a \|  \widehat{\rho}_\pm(\cdot, a) - \rho_{\pm}^*(\cdot, a, \widehat{q}_{\pm}) \| \right) - o_p(1) = -o_p(1), 
\end{align*} 
demonstrating the desired bound. 
\end{proof}

\begin{proof}[Proof of \cref{ex:erm}]
The $L_2$ convergence rate for a H\"older $\beta$-smooth functions in $d$ dimension is $O_P(n^{-1/(2+d/\beta)})$. Taking $e, \rho_\pm, q_\pm$ to be $\gamma_e$, $\gamma_\rho$, and $\gamma_q$-H\"older, we have that their convergence rates are $O_P(n^{-1/(2+d/\gamma_e)}), O_P(n^{-1/(2+d/\gamma_\rho)})$, and $O_P(n^{-1/(2+d/\gamma_q)})$ respectively. Thus, the $L_2$ conditional bias in \cref{thm:condneym} is bounded above by a term that is $O_p\big(n^{-2/(2+d/\gamma_q)} + n^{-(1/(2+d/\gamma_e)}\big)$. Applying \cref{thm:erm} with a $\beta$-smooth function class $\mathcal{F}$, we obtain the desired rate $O_p\big(n^{-1/(2+d/\beta)} + n^{-2/(2+d/\gamma_q)} + n^{-(1/(2+d/\gamma_e)  +1/(2+d/\gamma_\rho)}\big)$. The rest follows by algebraic manipulation.
\end{proof}

\begin{proof}[Proof of \cref{thm:drsmooth}]
We first derive:\allowdisplaybreaks
\begin{align*}
    \left| \widehat{\tau}^{\pm}(x) - \tau^\pm(x) \right| & = \left| \tilde{\tau}^{\pm}(x) - \tau^\pm(x) + \widehat{\tau}^{\pm}(x) - \tilde{\tau}^{\pm}(x) \right| \\
    & = \left| \tilde{\tau}^{\pm}(x) - \tau^\pm(x) + \frac{1}{n} \sum_{i=1}^n w_i(x) \left( \phi_\tau^{\pm}(Z_i, \widehat{\eta}) - \phi_\tau^{\pm}(Z_i, \eta^*)) \right) \right| \\
    & = \left| \tilde{\tau}^{\pm}(x) - \tau^\pm(x) + \frac{1}{n} \sum_{i=1}^n w_i(x) \left( \mathcal{E}_\tau^\pm(X_i; \widehat{\eta}) + \phi_\tau^{\pm}(Z_i, \widehat{\eta}) - \phi_\tau^{\pm}(Z_i, \eta^*) - \mathcal{E}_\tau^\pm(X_i; \widehat{\eta}) \right) \right| \\
    & \leq \left| \tilde{\tau}^{\pm}(x) - \tau^\pm(x) \right| + \left| \frac{1}{n} \sum_{i=1}^n w_i(x) \mathcal{E}_\tau^\pm(X_i; \widehat{\eta}) \right| + \left| \frac{1}{n} \sum_{i=1}^n w_i(x) \left( \phi_\tau^{\pm}(Z_i, \widehat{\eta}) - \phi_\tau^{\pm}(Z_i, \eta^*) - \mathcal{E}_\tau^\pm(X_i; \widehat{\eta}) \right) \right| 
\end{align*}

Since $\phi_\tau^{\pm}(Z_i, \widehat{\eta}) - \phi_\tau^{\pm}(Z_i, \eta) - \mathcal{E}_\tau^\pm(X_i; \widehat{\eta})$ is zero-meaned conditional on $X$ and nuisances (including weights), we can apply Chebyshev's inequality to randomness in $(A, Y) \mid X$ to obtain:
\begin{align*} 
    \left| \widehat{\tau}^{\pm}(x) - \tau^\pm(x) \right| & \leq \left| \tilde{\tau}^{\pm}(x) - \tau^\pm(x) \right| + \left| \frac{1}{n} \sum_{i=1}^n w_i(x) \mathcal{E}_\tau^\pm(X_i; \widehat{\eta}) \right| \\
    & + O_p\left( \| \widehat{\phi}_\tau^{\pm} - \phi_\tau^{\pm} - \mathcal{E}_\tau^\pm \|_{w^2} \frac{1}{n^2} \sum_{i=1}^n w_i(x)^2 \right)
\end{align*} 
Since $\mathcal{E}_\tau^\pm(X; \widehat{\eta}) = \E\left[ \widehat{\phi}_\tau^{\pm} - \phi_\tau^{\pm} \mid X \right]$, we can take advantage of the weighted $L^2$ norm and the weak law of large numbers to further bound $\| \widehat{\phi}_\tau^{\pm} - \phi_\tau^{\pm} - \mathcal{E}_\tau^\pm \|_{w^2} \leq \| \widehat{\phi}_\tau^{\pm} - \phi_\tau^{\pm} \|_{w^2} + o_p(1)$:
\begin{align*} 
    \left| \widehat{\tau}^{\pm}(x) - \tau^\pm(x) \right| & \leq \left| \tilde{\tau}^{\pm}(x) - \tau^\pm(x) \right| + b_n^\pm(x) + O_p\left( (\| \widehat{\phi}_\tau^{\pm} - \phi_\tau^{\pm} \|_{w^2} + o_p(1)) \left( \frac{1}{n^2} \sum_{i=1}^n w_i(x)^2 \right)^{1/2} \right),
\end{align*} 
which is the desired inequality. 
\end{proof}

\begin{proof}[Proof of \cref{cor:PointwiseLinearSmoother}]
By \citet{stone1977consistent} Theorem 1 and since $|\phi_\tau^{\pm}|$ is bounded, $\tilde{\tau}(x) \to^p \E[\phi_\tau^{\pm}(Z, \eta) \mid X=x] = \tau^\pm(x)$. 

For the second term, we use the \cref{thm:condneym} and the supremum assumptions to derive:
\begin{align*}
    | b_n^\pm(x)| & \leq \frac{1}{n} \sum_{i=1}^n |w_i(x)| \sup_x | \mathcal{E}_\tau^{\pm}(x; \widehat{\eta}) || \\
    & \lesssim \left( \sup_x |\widehat{e}(x) - e^*(x)| \right) \left( \sup_{x, a} |\widehat{\rho}_{\pm}(x, a) - \rho_{\pm}^*(x, a) | \right) + \sup_{x,a} ( \widehat{q}_\pm(x, a) - q_\pm^*(x, a))^2 \\
    & = o_p(1) 
\end{align*}

For the final term, the sup consistency implies $\| \widehat{\phi}_\tau^\pm - \phi_\tau^\pm \| = o_p(1)$. We also have:
\begin{align*}
    \frac{1}{n^2} \sum_{i=1}^n w_i(x)^2 & \leq \left( \frac{1}{n} \sum_{i=1}^n |w_i(x)| \right)^2 = O_p(1) 
\end{align*} 
So that $O_p\left( \frac{1}{n^2} \sum_{i=1}^n w_i(x)^2 \| \widehat{\phi}_\tau^\pm - \phi_\tau^\pm \|_{w^2} \right) = o_p(1)$.
\end{proof}

\begin{proof}[Proof of \cref{cor:LinearSmootherPointwiseValidity}]
If we define $\bar{\tau}^\pm(x)$ for the linear smoother estimate that uses $\bar{\eta}$ as first-stage nuisances, we can similarly argue that:
\begin{align*}
    \left| \widehat{\tau}^{\pm}(x) - \bar{\tau}^\pm(x) \right| & = \left| \tilde{\tau}^{\pm}(x) - \bar{\tau}^\pm(x) + \widehat{\tau}^{\pm}(x) - \tilde{\tau}^{\pm}(x) \right| \\
    & \leq \frac{1}{n} \sum_{i=1}^n |w_i(x)| \left| \mathcal{E}_\tau^{\pm}(X_i; \widehat{\eta}) - \mathcal{E}_\tau^{\pm}(X_i; \bar{\eta}) \right| \\
    & + O_p\left( \left( \| \phi_\tau^{\pm}(\cdot, \widehat{\eta}) - \phi_\tau^{\pm}(\cdot, \bar{\eta}) \|_{w^2} + o_p(1) \right) \left( \frac{1}{n^2} \sum_{i=1}^n w_i(x)^2 \right)^{1/2} \right) \\
    & = o_p(1) 
\end{align*}

By \citet{stone1977consistent} Theorem 1, $\bar{\tau}^\pm(x) \to^p \E[\phi_\tau^\pm(Z, \bar{\eta}) \mid X=x]$.

By \cref{thm:condneym}, $\pm \left( \E[\phi_\tau^\pm(Z, \bar{\eta}) \mid X=x] - \tau^{\pm}(x) \right) \geq 0$.

Therefore $\pm \left( \widehat{\tau}^{\pm}(x) - \tau^\pm(x) \right) \geq -o_p(1)$.
\end{proof}

\section{Detailed Algorithm}\label{sec:detailed-alg}

We present a more detailed version of the B-Learner pseudocode in \cref{alg:sharp-cate-full}.

\setcounter{algorithm}{0}
\begin{algorithm}[h]\caption{The B-Learner: Detailed}\label{alg:sharp-cate-full}
\begin{algorithmic}[1]
\INPUT Data $\{(X_i, A_i, Y_i): i\in \{1,...,n\}\}$, folds $K\geq 2$, sensitivity parameter $\Lambda \geq 1$, nuisance estimators, regression learner $\widehat{\EE}_n$.
 \FOR{$k\in \{1,...,K\}$} 
  \STATE Set $\mathcal{S}_k = \{(X_i, A_i, Y_i): i\neq k-1\;(\text{mod}\; K)\}$ 
  \\ Using $\mathcal{S}_k$:
  \STATE Learn outcome model:  $\widehat{\mu}^{(k)}(x,a) = \widehat{\EE}\left[Y|X=x,A=a\right]$
  \STATE Learn propensity model: $\widehat e^{(k)}(x)= \widehat{p}(A=1|X=x)$
  \STATE Learn conditional outcome quantile models:\\ $\widehat{q}^{(k)}_+(x,a)=\widehat{\inf}\{\beta: F(\beta\mid X=x, A=a)\geq \Lambda / (\Lambda+1) \}$  \\ $\widehat{q}^{(k)}_-(x,a)=\widehat{\inf}\{\beta: F(\beta\mid X=x, A=a)\geq 1 / (\Lambda+1) \}$ 
  \STATE Learn conditional value at risk models: \\
  $\widehat{\text{CVaR}}^{(k)}_\pm(x,a) = $  $\widehat{q}^{(k)}_\pm(x,a)+ (\Lambda+1)\widehat{\EE}\left[\{Y-\widehat{q}^{(k)}_\pm(x,a)\}_{\pm} \mid X=x,A=a\right]$ \\
   \STATE  Set $\widehat{\rho}_{\pm}^{(k)}(x,a) = \Lambda^{-1} \widehat{\mu}^{(k)}(x,a) + (1-\Lambda^{-1})\widehat{\text{CVaR}}^{(k)}_{\pm}(x,a)$
    \FOR{$i= k-1\;(\text{mod}\; K)$} 
        \STATE Set $R_{\pm,i} = \Lambda^{-1}Y_i + (1-\Lambda^{-1}) \left( \widehat{q}^{(k)}_\pm(X_i,A_i)+ \frac{1}{1-\alpha} \{Y_i-\widehat{q}^{(k)}_\pm(X_i,A_i)\}_{\pm} \right) $
        \STATE Set pseudo-outcomes for $Y^\pm(X, 1)$: \\
$\widehat{\phi}^{\pm}_{1,i} =  A_i Y_i + (1-A_i)\widehat{\rho}_{\pm}^{(k)}(X_i, 1) 
         \ts+\frac{\left(1-\widehat{e}^{(k)}(X_i)\right)A_i}{\widehat{e}^{(k)}(X_i)}\cdot \left(R_{\pm,i}-\widehat{\rho}^{(k)}_{\pm}(X_i, 1)\right)$
        
         \STATE   Set pseudo-outcomes for $Y^\pm(X, 0)$:\\ 
         $\widehat{\phi}^{\pm}_{0,i} =  (1-A_i)Y_i + A_i\widehat{\rho}_{\pm}(X_i, 1) 
         + \ts\frac{\widehat{e}^{(k)}(X_i)(1-A_i)}{\left(1-\widehat{e}^{(k)}(X_i)\right)}\cdot \left(R_{\pm,i}-\widehat{\rho}^{(k)}_{\pm}(X_i, 0)\right)$
        \STATE Set pseudo-outcomes for CATE: \\ $\widehat{\phi}^{\pm}_{\tau, i}=\widehat{\phi}^{\pm}_{1,i} - \widehat{\phi}^{\mp}_{0,i}$
\ENDFOR
\ENDFOR
\STATE Create datasets $\mathcal{T}^{\pm} = \{(X_i, \widehat{\phi}^{\pm}_{\tau, i})\}$
\STATE Learn upper- and lower- bound functions  $\widehat{\tau}^{\pm}(x) = \widehat{\EE}_n\left[\widehat{\phi}^{\pm}_{\tau}\mid X=x\right] $ from the datasets  $\mathcal{T}^{\pm}$
\OUTPUT $\widehat{\tau}^{\pm}$
\end{algorithmic}
\end{algorithm}

\section{Additional Experimental Details}\label{sec:add-exp-results}

The replication code for all simulations is distributed under an MIT license.

\subsection{Simulated Data}

The results in Section 8 were obtained using an Amazon Web Services instance with 32 vCPUs and 64 GiB of RAM. For the Random Forest (RF) models, we use the \texttt{RandomForestRegressor} model from \texttt{scikit-learn}. For Gaussian Kernels (GK), we use \texttt{RBF} (radial basis function) method from \texttt{scikit-learn}. Finally, for the Bayesian Neural Networks (NN) we use several functions from the \texttt{PyTorch} package. The quantile estimators use weights from the nuisance regressors when RFs or GKs are used or are calculated from the sampled outcome distributions when NNs are used. We include the hyperparameters for the different models used with the synthetic data in \cref{tab:sim_hp}.

\begin{center}
\captionof{table}{Hyperparameters for model choices in synthetic data experiments.}
\vspace{0.5em}
\begin{tabular}{l l l } 
 \hline
 Model & Hyperparameter & Value  \\ [0.5ex] 
 \hline\hline
 Random Forest (\texttt{scikit-learn}) & max\_depth & 6 \\
 & min\_samples\_leaf & 0.05 \\
\hline
RBF (\texttt{scikit-learn}) & length\_scale & $0.9\times n^{-\frac{1}{4+d}}$\\
\hline
Neural Network (\texttt{PyTorch}) & hidden units & 100\\
& network depth & 4\\
& negative slope & 0.3\\
& dropout rate & 0.2\\
& batch size & 50\\
& learning rate & $5e$-4\\
\hline
\end{tabular}
\label{tab:sim_hp}
\end{center}

\subsection{IHDP Dataset}\label{sec:datasets}
We use \citet{jesson2021quantifying}'s hidden confounding version of the Infant Health and Development Program (IHDP) that was introduced by \citet{hill2011bayesian}. The data comes from an experiment that targeted “low-birth-weight, premature infants, and provided the treatment group with both intensive high-quality child care and home visits from a trained provider” \citep{hill2011bayesian}. For the purpose of simulating an observational study, \citet{hill2011bayesian} generates simulated outcomes using the following features: measurements on the child–birth weight, head circumference, weeks born preterm, birth order, firstborn, neonatal health index, sex, twin status–as well as behaviors engaged in during pregnancy–smoked cigarettes, drank alcohol, took drugs–and measurements on the mother at the time she gave birth–age, marital status, educational attainment (did not graduate from high school, graduated from high school, attended some college but did not graduate, graduated from college), whether she worked during pregnancy, whether she received prenatal care, and the site (8 total) in which the family resided at the start of the intervention. A non-random portion of the treatment group, the children of non-white mothers, are excluded from the study in order to mimic confounding in an otherwise randomized trial.
Covariates consist of 6 continuous variables and 19 binary variables. 
We use the covariate descriptions from \citet{jesson2021quantifying} which we replicate in \cref{tab:ihdp_covariates} for completeness. The dataset consists of 747 samples, of which 139 are in the treatment group.

\begin{table}[]
    \centering
    \caption{\textbf{IHDP Covariates} Binary covariates $x_{9}-\mathrm{x}_{18}$ are attributes of the mother. Mother's education level ``College" indicated by covariates $\mathrm{x}_{10}-\mathrm{x}_{12}$ all zero. Site 8 indicated by covariates $\mathrm{x}_{19}-\mathrm{x}_{25}$ all zero. We show the frequency of occurrence for each binary covariate $p(\mathrm{x} = 1)$, as well as the adjusted mutual information $I(\mathrm{x}; \ti)$ between the binary covariate and the treatment variable.}
    \vspace{0.5em}
    \begin{tabular}{ll|llcc}
        \toprule
        \multicolumn{2}{l|}{Continuous} & \multicolumn{4}{l}{Binary}  \\
        Covariate           & Description               & Covariate         & Description                       & $I(\mathrm{x}; \ti)$  & $p(\mathrm{x} = 1)$   \\
        \midrule
        $\mathrm{x}_{1}$    & birth weight               & $\mathrm{x}_{7}$  & child's gender (female=1)         & 0.00                  & 0.51                  \\
        $\mathrm{x}_{2}$    & head circumference        & $\mathrm{x}_{8}$  & is child a twin                   & 0.00                  & 0.09                  \\
        $\mathrm{x}_{3}$    & number of weeks pre-term  & $\mathrm{x}_{9}$  & married when child born           & \textbf{0.02}         & \textbf{0.52}         \\
        $\mathrm{x}_{4}$    & birth order               & $\mathrm{x}_{10}$ & left High School                  & 0.00                  & 0.36                  \\
        $\mathrm{x}_{5}$    &``neo-natal health index"  & $\mathrm{x}_{11}$ & completed High School             & 0.00                  & 0.27                  \\
        $\mathrm{x}_{6}$    & mom's age                 & $\mathrm{x}_{12}$ & some College                      & 0.00                  & 0.22                  \\
                            &                           & $\mathrm{x}_{13}$ & child is first born               & 0.00                  & 0.36                  \\
                            &                           & $\mathrm{x}_{14}$ & smoked cigarettes when pregnant   & \textbf{0.01}         & \textbf{0.48}         \\
                            &                           & $\mathrm{x}_{15}$ & consumed alcohol when pregnant    & 0.00                  & 0.14                  \\
                            &                           & $\mathrm{x}_{16}$ & used drugs when pregnant          & 0.00                  & 0.96                  \\
                            &                           & $\mathrm{x}_{17}$ & worked during pregnancy           & \textbf{0.01}         & \textbf{0.59}         \\
                            &                           & $\mathrm{x}_{18}$ & received any prenatal care        & \textbf{0.01}         & 0.96                  \\
                            &                           & $\mathrm{x}_{19}$ & site 1                            & 0.00                  & 0.14                  \\
                            &                           & $\mathrm{x}_{20}$ & site 2                            & \textbf{0.01}         & 0.14                  \\
                            &                           & $\mathrm{x}_{21}$ & site 3                            & 0.00                  & 0.16                  \\
                            &                           & $\mathrm{x}_{22}$ & site 4                            & \textbf{0.01}         & 0.08                  \\
                            &                           & $\mathrm{x}_{23}$ & site 5                            & \textbf{0.02}         & 0.07                  \\
                            &                           & $\mathrm{x}_{24}$ & site 6                            & \textbf{0.01}         & 0.13                  \\
                            &                           & $\mathrm{x}_{25}$ & site 7                            & \textbf{0.02}         & 0.16                  \\
        \bottomrule
    \end{tabular}
    \label{tab:ihdp_covariates}
\end{table}

\citet{jesson2021quantifying} create the Hidden Confounding of IHDP by hiding the covariate $\mathrm{x}_{9}$ from models during training, however, the causal model depends on it for the data generation. Following is the data generation process of the Hidden Confounding version of response surface B \cite{hill2011bayesian}, we restate the data generation process from \citet{jesson2021quantifying}:
\begin{subequations}
    \begin{align}
        \ui &\coloneqq N_{\ui}, \\
        \x &\coloneqq N_{\x}, \\
        \ti &\coloneqq N_{\ti}, \\
        \y &\coloneqq (\ti - 1) ( \exp(\beta_{\x}(\x + \mathbf{w}) + \beta_{\ui}(\ui + 0.5)) + N_{\Yzero}) + \ti (\beta_{\x}\x + \beta_{\ui}\ui - \omega^s + N_{\Yone})),
    \end{align}
\end{subequations}
where $(N_{\ui}, N_{\x}, N_{\ti}) \sim p_{\D}(\mathrm{x}_{9}, \{\mathrm{x}_{1}, \dots, \mathrm{x}_{8}, \mathrm{x}_{10}, \dots, \mathrm{x}_{25} \}, \ti)$, $N_{\Yzero} \sim \mathcal{N}(0, 1)$, and $N_{\Yone} \sim \mathcal{N}(0, 1)$.
The coefficient $\beta_{\ui}$ is randomly sampled from $( 0.1, 0.2, 0.3, 0.4, 0.5)$ with probabilities $(0.2, 0.2, 0.2, 0.2, 0.2)$, $\beta_{\x}$ is a vector of randomly sampled values $( 0.0, 0.1, 0.2, 0.3, 0.4)$ with probabilities $(0.6, 0.1, 0.1, 0.1, 0.1)$, $w$ is a vector with all the coordinates equals $0.5$, where $\omega^s$ was chosen as in \citet{hill2011bayesian}: "for the s$^{\text{th}}$ simulation, it was chosen in the overlap setting, where we estimate the effect of the treatment on the treated, such that CATT equals 4; similarly it was chosen in the incomplete setting, where we estimate the effect of the treatment on the controls so that CATC equals 4". 

Following \citet{jesson2021quantifying}'s Hidden Confounding experiment, we generate 400 realizations of the IHDP dataset, such that the seed for each realization is the corresponding index of the realization, where the indices are $0,1,....,400$. Each realization is split into training ($n=470$), validation ($n=202$), and test ($n=75$) subsets. For the B-Learner with NNs, we use the same models and hyperparameters used by \textit{Quince} in \citet{jesson2021quantifying}. For the B-Learner with RF base estimators, we use the \texttt{RandomForestRegressor} from \texttt{scikit-learn} and \texttt{ForestRegressor} from \texttt{econml.grf} where we control for forest growth only through the max\_depth ($=6$) and min\_samples\_leaf ($=0.01$) parameters. As for \textit{Kernel Sensitivity} and \textit{Quince}, to replicate the results from \citet{jesson2021quantifying}, we use the same models and hyperparameters they used for the Hidden Confounding IHDP experiment. \textit{Note:} we exclude $5$ of the $400$ IHDP trials from the original analysis due to poor data quality (e.g. low overlap) that affects the NN training. These issues seem to be mitigated by ensembling which is why they do not pose a problem for the experiments in \citet{jesson2021quantifying}. We will perform a comparison of the ensembled methods in a future iteration of this work.    

\subsection{401(k) Eligibility Study}

The dataset includes 9{,}915 observations with 9  covariates such as age, income, education, family size, marital status, IRA participation, etc. We describe the features of the 401(k) dataset in \cref{tab:401k_feats}. In order to replicate the CATEs obtained by \cite{chernozhukov2018double}, we use the same models (\texttt{RandomForestRegressor} and \texttt{RandomForestClassifier} from \texttt{scikit-learn}) and hyperparameters (n\_estimators = 100, max\_depth = 7, max\_features = 3, min\_samples\_leaf = 10) for our nuisance estimators and second stage models.

\begin{center}
\captionof{table}{Features of 401(k) dataset.}
\vspace{0.5em}
\begin{tabular}{l l l } 
 \hline
 Name & Description & Type  \\ [0.5ex] 
 \hline\hline
age & age & continuous  \\
inc & income & continuous \\
educ & years of completed education & continuous \\
fsize & family size & continuous \\
marr & marital status & binary \\
two\_earn & whether dual-earning household & binary\\
db & defined benefit pension status & binary \\
pira & IRA participation & binary \\
hown & home ownership & binary\\
e401 & 401 (k) eligibility & binary \\
net\_tfa & net financial assets & continuous\\
\hline
\end{tabular}
\label{tab:401k_feats}
\end{center}

\end{document}